\documentclass[letterpaper]{article} % DO NOT CHANGE THIS
\usepackage{aaai25}  % DO NOT CHANGE THIS
\usepackage{times}  % DO NOT CHANGE THIS
\usepackage{helvet}  % DO NOT CHANGE THIS
\usepackage{courier}  % DO NOT CHANGE THIS
\usepackage[hyphens]{url}  % DO NOT CHANGE THIS
\usepackage{graphicx} % DO NOT CHANGE THIS
\usepackage{natbib}  % DO NOT CHANGE THIS AND DO NOT ADD ANY OPTIONS TO IT
\usepackage{caption} % DO NOT CHANGE THIS AND DO NOT ADD ANY OPTIONS TO IT
\usepackage{algorithm}
\usepackage{algorithmic}
\usepackage{amssymb}
\usepackage{amsmath}
\usepackage{multirow}
\usepackage{booktabs}
\usepackage{colortbl}
\usepackage[table,xcdraw]{xcolor}
\usepackage{float}

\usepackage{newfloat}
\usepackage{listings}
\DeclareCaptionStyle{ruled}{labelfont=normalfont,labelsep=colon,strut=off} % DO NOT CHANGE THIS
\floatstyle{ruled}
\newfloat{listing}{tb}{lst}{}
\floatname{listing}{Listing}
\title{
Attribution Analysis Meets Model Editing: Advancing Knowledge Correction in Vision Language Models with~\emph{VisEdit}
}
\author {
    Qizhou Chen\textsuperscript{\rm 1},
    Taolin Zhang\textsuperscript{\rm 2},
    Chengyu Wang\textsuperscript{\rm 3}\footnotemark[1],
    Xiaofeng He\textsuperscript{\rm 1}\thanks{Co-corresponding Authors},
    Dakan Wang\textsuperscript{\rm 4},
    Tingting Liu\textsuperscript{\rm 2} 
}
\affiliations {
    \textsuperscript{\rm 1} East China Normal University, Shanghai, China\\
    \textsuperscript{\rm 2} Alibaba Group, Hangzhou, China\\
    \textsuperscript{\rm 3} Alibaba Cloud Computing, Hangzhou, China\\
    \textsuperscript{\rm 4} Exacity Inc., Shanghai, China\\
    52265901009@stu.ecnu.edu.cn,  chengyu.wcy@alibaba-inc.com,
    hexf@cs.ecnu.edu.cn
}

\usepackage{bibentry}
\begin{document}

\maketitle

\begin{abstract}
Model editing aims to correct outdated or erroneous knowledge in large models without costly retraining. Recent research discovered that the mid-layer representation of the subject's final token in a prompt has a strong influence on factual predictions, and developed Large Language Model (LLM) editing techniques based on this observation. However, for Vision-LLMs (VLLMs), how visual representations impact the predictions from a decoder-only language model remains largely unexplored. To the best of our knowledge, model editing for VLLMs has not been extensively studied in the literature. In this work, we employ the contribution allocation and noise perturbation methods to measure the contributions of visual representations for token predictions. Our attribution analysis shows that visual representations in mid-to-later layers that are highly relevant to the prompt contribute significantly to predictions. Based on these insights, we propose \emph{VisEdit}, a novel model editor for VLLMs that effectively corrects knowledge by editing intermediate visual representations in regions important to the edit prompt. We evaluated ~\emph{VisEdit} using multiple VLLM backbones and public VLLM editing benchmark datasets. The results show the superiority of~\emph{VisEdit} over the strong baselines adapted from existing state-of-the-art editors for LLMs.

\end{abstract}

\section{Introduction}
With increasing number of LLM applications \cite{DBLP:journals/corr/abs-2302-13971,DBLP:journals/fi/RoumeliotisT23, DBLP:conf/iclr/ZengLDWL0YXZXTM23}, there's a rising demand for updating the static knowledge inside the LLM using model editing techniques \cite{KnowledgeEditor, MEND, ZJUEditSurvey2023}.  
It aims to efficiently correct knowledge within LLMs without the necessity for retraining. 
This technology plays a key role in eliminating illusions \cite{DBLP:journals/corr/abs-2401-06855,DBLP:journals/corr/abs-2310-16045}, reducing bias \cite{DBLP:journals/corr/abs-2310-18913,DBLP:conf/emnlp/AkyurekPKW23}, and protecting privacy \cite{DBLP:journals/corr/abs-2309-11852, DBLP:conf/emnlp/WuLXDW0X23} for LLMs.

Recent model editing works primarily focus on the text modality only. They can be broadly categorized into three types: 
(1) Methods which directly modify parameters \cite{MEND,ROME, MEMIT,WILKE} typically locate and modify the MLP weights important for token prediction.
(2) Methods which add extra modules \cite{T-Patcher,GRACE,MELO} make response correction by constructing additional bypass modules for the model. 
(3) Methods which use prefix editing \cite{InContextKnowledgeEdit,LTE,RECIPE} aim to make the model follow editing instructions inserted before the input text.

Compared to single-modality cases, editing Vision-LLMs (VLLMs) that incorporate both visual and text inputs poses unique challenges, and have not been widely explored.
\citet{MMEdit} attempted to adapt LLM editors to VLLMs and established quantitative metrics for VLLM editing. Their experiments suggest that LLM editors are not very suitable for VLLMs. 
They hypothesize that the cause of response errors may not only stem from the weights of the LLM decoder but also from the interactions between the two modalities. 
In exploring the connections between text and visual modalities, some efforts \cite{DBLP:conf/iccvw/SchwettmannCKB023,DBLP:journals/corr/abs-2311-07470} focus on identifying visual neurons within the LLM decoder. 
\citet{DBLP:journals/corr/abs-2406-04236} reveal that a few final output representations of the visual encoder significantly contribute to visual constraints in the prompt at the LLM's first layer. 
Nonetheless, it is still unknown how visual representations affect the final predictions within the whole computation graph of VLLMs. Understanding this can better elucidate the response generation process of VLLMs and benefit VLLM editing.

To address these challenges, we first analyze the impact of visual representations on token predictions in VLLMs based on contribution allocation and noise perturbation methods. Based on the analysis results, we introduce a novel VLLM editor named~\emph{VisEdit}. Our works are summarized below.

\begin{figure*}[t]
\centering
\includegraphics[width=1\textwidth]{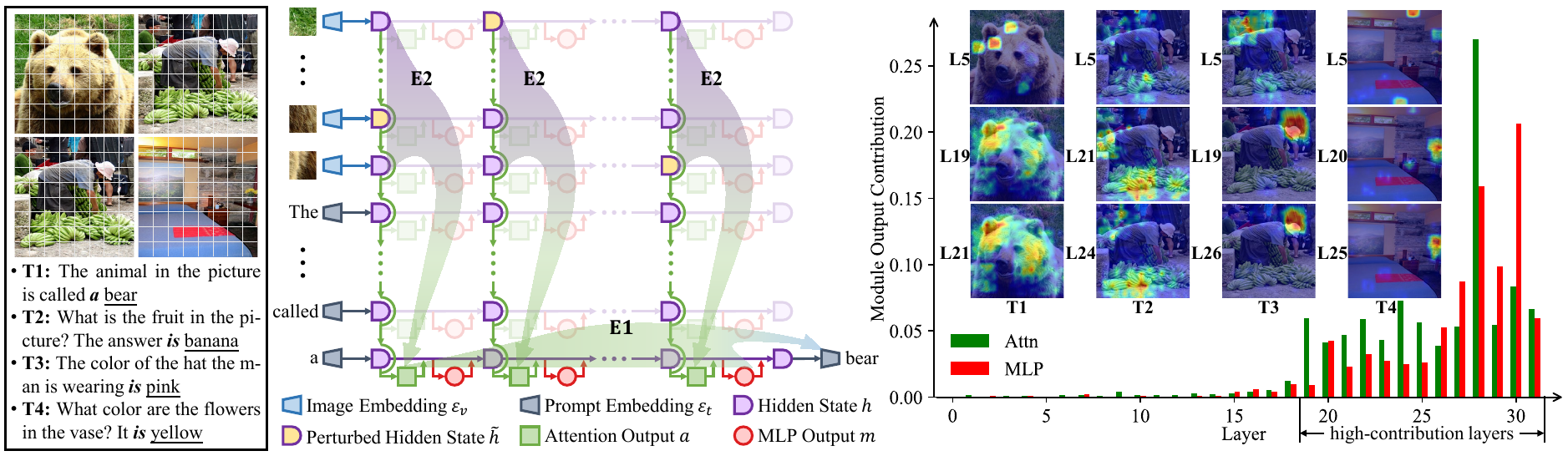} 
\caption{
Attribution analysis for LLaVA-V1.5 \cite{llava}.
\textbf{E1:} Measuring contributions of the attention and MLP outputs at each layer to the prediction of a key token. Average results on E-VQA \cite{MMEdit} dataset are displayed in the bar chart.
\textbf{E2:} Measuring the contributions of visual representations to the attention module outputs. 
The results for four samples are visualized in the heatmaps, where red indicates higher contributions and blue indicates lower.
\textbf{T*} and \textbf{L*} respectively indicate the test sample index and the layer index selected by the visual representations attribution analysis.
In each sample, the italicized bold text and the underlined text respectively represent the last token used for prediction and the key token to be predicted.
}
\label{img_attribution}
\vspace{-1em}
\end{figure*}

\noindent\textbf{Visual Representation Attribution:}
The middle part of Figure \ref{img_attribution} illustrates the dependencies between hidden states in the computation graph of a typical VLLM. 
% To measure each visual representation's contribution to the key token (e.g., 'bear'), we first assess the contribution of each module's output, thereby reducing interference between representations.
% Since the logits are obtained by cumulatively adding the output of the attention and MLP modules across layers, each output directly contributes to the logits.
% Therefore, we measure each module's contribution to the key token prediction by mapping its output to the vocabulary probability space.
Firstly, we measure the contributions of each layer's MLP and attention modules to the key token (e.g., ``bear'') prediction by mapping each module's output to the predicted probability of the key token.
Subsequently, we employ a noise perturbation-based attribution method \cite{DBLP:conf/kdd/Ribeiro0G16, DBLP:conf/iccv/FongV17, ROME} to assess each visual representation's contribution to the subsequent layer’s attention module output. 
% Specifically, we introduce noise into each visual representation and then calculate a new attention output together with the remaining clean representations. 
For one specific layer, we perturb one visual hidden state while fixing the others, and then compute a new attention output.
A smaller similarity between the new attention output and the original one indicates the change in the selected visual representation affects the attention module output more significantly, or makes a greater contribution in simple terms.
The brief results of these two steps are shown on the right side of Figure \ref{img_attribution}, which indicate: 
\begin{itemize}
% \item Outputs of deep model layers contribute more significantly to the key token . 
\item Outputs of deep model layers contribute more significantly to the key token compared with shallow layers.
\item For high-contribution layers, the attention outputs are mostly influenced by the visual region of the objects mentioned in the prompts.
% the spatial distribution of the visual representation's contribution to its attention module output is highly correlated with the object mentioned in the prompt.

\end{itemize}
Based on the observations, we hypothesize that:~\emph{In VLLMs, the early layers tend to aggregate the information queried in the prompt at the last token. This setup allows deeper layers to extract information from visual representations of key regions, facilitating the generation of responses.}

\noindent\textbf{The Proposed VLLM Editor~\emph{VisEdit}:} 
According to the above results, we design an effective VLLM editor. Based on the first attribution experiment, we place a trainable visual representation adaptor before the high-contribution layer. 
% The adaptor applies cross-attention to infuse editing information into the visual representations, thereby correcting model responses through subsequent vision-language interactions.
% Based on the second experiment, we also introduce an Influence Mapper (IM) to decide the fusing intensity of the edit signal.
% We regularize the IM to attend to the key visual regions relevant to the edit prompt.
% It allows the editor to focus on the visual representation that are crucial for modifying key responses, thus enhancing the efficacy of editing while minimizing the impact on irrelevant visual representations.
The adaptor applies cross-attention to infuse information of the edit sample into the visual representations of a given input sample. To ensure that the adaptation of visual representations is applied to the most important regions, based on the second experiment, we introduce an influence mapper module to identify the key visual regions most relevant to the edit prompt. In this way, the editor will focus on the visual representations that are crucial for modifying key responses, thereby enhancing the efficacy of editing while leaving irrelevant visual representations mostly untouched.
We conduct editing experiments on three typical VLLMs, including BLIP2-OPT (2.7B)~\cite{BLIP2}, MiniGPT-4 (7B)~\cite{MiniGPT-4}, and LLaVA-V1.5 (7B)~\cite{llava}, with two VLLM editing datasets, E-VQA and E-IC \cite{MMEdit}. The experimental results demonstrate that our method excels in reliability, text/modal generality, and text/modal locality metrics, which follow standard evaluation protocols of the benchmark \cite{MMEdit}. Furthermore, comprehensive exploratory experiments have validated the effectiveness of our module designs.
Source codes are available at \url{https://github.com/qizhou000/VisEdit}.

\section{Related Works}
\subsection{Vision Language Models}
% The early vision-language works primarily focusing on single tasks and using limited supervised datasets for training \cite{DBLP:journals/corr/KirosSZ14, DBLP:conf/cvpr/VinyalsTBE15, DBLP:conf/cvpr/KarpathyL15, DBLP:conf/iccv/AntolALMBZP15}.
% % With the advancement of LLMs \cite{DBLP:journals/corr/abs-2302-13971,DBLP:journals/fi/RoumeliotisT23, DBLP:conf/iclr/ZengLDWL0YXZXTM23} facilitated by Transformer \cite{DBLP:conf/nips/VaswaniSPUJGKP17}, 
% Then, the vision transformer (ViT) \cite{DBLP:conf/iclr/DosovitskiyB0WZ21} enabled the transformation from images to sequences, facilitating the vision-language pre-training \cite{DBLP:conf/icml/RadfordKHRGASAM21} changed the previous paradigm of purely supervised learning for vision-language interaction.
% and significantly enhanced the semantic understanding capability of LLMs for visual modalities.

Currently, VLLMs benefited from Vision Transformer (ViT) \cite{DBLP:conf/iclr/DosovitskiyB0WZ21} and vision-language pre-training \cite{DBLP:conf/icml/RadfordKHRGASAM21} can be generally categorized into two types \cite{DBLP:journals/corr/abs-2405-17927}, Modal Deep Fusion (MDF) and Modal Early Fusion (MEF). 
MDF fuses visual and language modalities within the internal layers of LLMs using cross-modal attention \cite{DBLP:conf/nips/LuBPL19,DBLP:conf/nips/AlayracDLMBHLMM22, DBLP:journals/corr/abs-2401-16420}.
% allowing for fine-grained control over the flow of modality information \cite{}. 
MEF integrates visual modalities before feeding them into the LLM by training specific encoding modules \cite{DBLP:journals/corr/abs-2311-04257,DBLP:journals/corr/abs-2308-12966, BLIP2, MiniGPT-4, llava}.
For example, BLIP2 \cite{BLIP2} and MiniGPT-4 \cite{MiniGPT-4} compress visual representations through a Q-Former, while LLaVA \cite{llava} directly trains an MLP layer after the visual encoder. 
Due to its modular architecture, MEF offers greater extensibility and has garnered more attention compared to MDF \cite{DBLP:journals/corr/abs-2405-17927}. Therefore, this paper primarily focuses on model editing for MEF-based VLLMs.

\subsection{Attribution Analysis}
Attribution analysis enhances the interpretability of deep learning models \cite{DBLP:journals/csur/GuidottiMRTGP19}, with this section focusing on perturbation-based methods. For instance, \citet{DBLP:conf/kdd/Ribeiro0G16, DBLP:conf/iccv/FongV17} use random masking or perturbation to assess contributions of image parts. Combining feature perturbation with causal mediation analysis \cite{DBLP:conf/uai/Pearl01, DBLP:conf/nips/VigGBQNSS20}, \citet{ROME} identify LLM layers that significantly influence responses. For visual attribution in VLLMs, methods like integrated gradients \cite{DBLP:conf/icml/SundararajanTY17} are used to locate neurons recognizing specific visual concepts \cite{DBLP:conf/iccvw/SchwettmannCKB023, DBLP:journals/corr/abs-2311-07470}. \citet{DBLP:journals/corr/abs-2406-04236} demonstrate that the last few visual representations significantly influence the visual constraints of the prompt at the first model layer. However, these works do not consider how the spatial semantics of visual representations influence VLLM prediction.

\subsection{Model Editing}
\noindent\textbf{Model Editing for LLMs:}
Model editing methods can be categorized into three types: modifying parameters, adding extra modules, and editing prefix. 
In modifying parameter-based methods, KE \cite{KnowledgeEditor} and MEND \cite{MEND} generate model weight offsets according to edit signals by training an auxiliary network. 
ROME \cite{ROME} and MEMIT \cite{MEMIT} utilize causal mediation analysis \cite{DBLP:conf/uai/Pearl01, DBLP:conf/nips/VigGBQNSS20} to locate FFN weights that have a significant causal effect on the response of LLMs, and then edit them.
% the weights through back-propagation. 
In adding extra modules-based methods, SERAC \cite{SERAC} trains a counterfactual model to respond to redirected queries related to the edit sample. 
TP \cite{T-Patcher} trains a piece of knowledge to be edited as an additional neuron. 
% GRACE \cite{GRACE} remaps the middle representations of the query if its distance to the editing representation is below a certain threshold.
GRACE \cite{GRACE} maps the intermediate layer input of the query to the output in editing space if the distance between their representations is below a threshold. 
MELO \cite{MELO} adjusts LLM weights by adding query-related editing matrices, retrieved through a mechanism similar to GRACE.
In editing prefix-based methods, IKE \cite{InContextKnowledgeEdit} 
% constructs demonstrations and 
uses in-context learning to modify LLMs' responses. 
LTE \cite{LTE} fine-tunes LLMs to follow editing instructions.
RECIPE \cite{RECIPE} trains a continuous prompt generator to find the shortest editing prefix.
Among them, KE and IKE explored the feasibility of related technologies in single editing. SERAC, GRACE, LTE, and RECIPE, extended single editing to multiple by incorporating retrieval mechanisms. To avoid redundant discussions on editing retrieval, we primarily explore the contributing mechanisms of visual representations in VLLM to response generation to inspire the design of VLLM single editing.

\noindent\textbf{Model Editing for VLLMs:}
To date, editing methods specific to VLLMs have not been widely studied. 
\citet{MMEdit} first attempts to migrate LLM editors to edit VLLM and constructs VLLM editing datasets along with corresponding evaluation metrics.
\citet{DBLP:journals/corr/abs-2402-14835} and \citet{DBLP:journals/corr/abs-2406-13219} expand the VLLM editing datasets related to entity knowledge and modality consistency, which are not yet open-sourced. 
% Based on the attribution results, we propose a VLLM-specific editor to provide a novel perspective to promote further research on the generation and correction of VLLM responses.
To address the gap in VLLM editing research, we analyze how spatial semantics of visual representations influence VLLM responses and propose a VLLM-specific editor to encourage further research on the generation and correction of VLLMs.

\section{Attribution Analysis}
\label{sec_attribution_analysis}
% To mitigate interference between representations and simplify the attribution process, we first evaluate the contribution of module outputs in each layer to the predictions, thereby allocating the assessment of each visual representation's contribution to its respective layer. Subsequently, we employ a noise perturbation method \cite{DBLP:conf/kdd/Ribeiro0G16,DBLP:conf/iccv/FongV17, ROME} to assess the contribution of visual representations at each layer to the attention output. 
We evaluate the impact of the visual representations on token prediction in two steps. First, we compute the contribution of each layer's module output to the predicted key token. Then, for a specific layer, we employ a noise perturbation method to evaluate how changes in a visual hidden state (synonymous with representation in the context) affect the attention module output.
Figure \ref{img_attribution} displays the data flow of VLLM response generation and the results of two attribution experiments.
For more attribution analysis results, please refer to Appendix \ref{sec_more_attribution_analysis}.

Given a VLLM $f_\theta:\mathcal{X}_v\times\mathcal{X}_t\mapsto \mathcal{O}$ that maps an image-prompt pair $(x_v, x_t)$ into a text response $o=f_\theta(x_v, x_t)$, we set $\hat{f}_\theta:\mathcal{E}_v\times\mathcal{E}_t\mapsto \mathcal{Y}$ as the transformer in $f_\theta$ that maps an embedding sequence $\boldsymbol{\varepsilon} = \boldsymbol{\varepsilon}_v\oplus \boldsymbol{\varepsilon}_t \in \mathcal{R}^{N\times d_h}$ to a probability distribution $y\in \mathcal{Y}\subset \mathcal{R}^{|\mathcal{V}|}$ predicting the next token based on vocabulary $\mathcal{V}$, where $\boldsymbol{\varepsilon}_v\in \mathcal{E}_v\subset \mathcal{R}^{N_v\times d_h}$ is the image embedding and $\boldsymbol{\varepsilon}_t\in \mathcal{E}_t\subset \mathcal{R}^{N_t\times d_h}$ is the prompt embedding. 
$N_v, N_t, N$ are the length of visual, text, and complete embeddings, respectively. 
$d_h$ is the middle dimension of $\hat{f}_\theta$ and $\oplus$ denotes concatenation. 
In our attribution experiments, we represent each sample as a tuple $(\boldsymbol{\varepsilon}_v, \boldsymbol{\varepsilon}_t, o^*)$, containing the image embedding $\boldsymbol{\varepsilon}_v$, the prompt embedding $\boldsymbol{\varepsilon}_t$, and the key token $o^*$ to be predicted. 
The symbols and notations used in this paper are provided in Appendix \ref{appendix_notations}.
 
\subsection{Module Output Attribution}
In a transformer $\hat{f}_\theta$, each hidden representation $h_n^l\in \mathcal{R}^{d_h}$ of $n$-th token at $l$-th layer can be obtained from \cite{DBLP:conf/nips/VaswaniSPUJGKP17}: 
\begin{gather}
    h_n^l = h_n^{l-1} + a_n^l + m_n^l, l\in\{1,..., L\}, n \in \{1,...,N\}
    \label{eq_transformer_formulation}
\end{gather}
where $h^0$ equals $\boldsymbol{\varepsilon}$, and $a_n^l = \text{Attn}_l(h_1^{l-1},...,h_n^{l-1})$, $m_n^l = \text{MLP}_l(h_n^{l-1}+a_n^l)$ are respectively the outputs of attention and MLP modules. $L$ is the layer count of $\hat{f}_\theta$. Given an input $(\boldsymbol{\varepsilon}_v, \boldsymbol{\varepsilon}_t)$, the predicted probability distribution of the next token is
$p = \delta(h_N^LW_{\mathcal{V}})$,
where $\delta$ is the softmax function, and $W_{\mathcal{V}}\in \mathcal{R}^{{d_h}\times |\mathcal{V}|}$ is the matrix mapping hidden states into logits in vocabulary space $\mathcal{V}$. 
% Since the last hidden state is the accumulative sum of the module outputs, i.e., 
Unrolling Equation \ref{eq_transformer_formulation}, we have 
$h_N^L = h_N^0+\sum_{l=1}^L\left(a_N^l + m_N^l\right)$. 
Multiplying both sides by $W_\mathcal{V}$ gives:
\begin{align}
    h_N^LW_{\mathcal{V}} = h_N^0W_{\mathcal{V}}+\sum\limits_{l=1}^L\left(a_N^lW_{\mathcal{V}} + m_N^lW_{\mathcal{V}}\right)
\end{align}
% It shows that the output of each module additively supports a certain token prediction distribution, respectively.
% It can be observed that the output of each module contributes additively to the final token prediction.
It shows that the token prediction distribution depends on the sum of the outputs from all attention and MLP modules across layers.
Therefore, we quantify the contribution of a module output $r$, which can be either $a_N^l$ or $m_N^l$, to the prediction of the key token $o^*$, as below:
\begin{gather}
    C_{o^*}(r) = \sqrt{C_{o^*}^p(r) \cdot C_{o^*}^v(r)}\\
    C_{o^*}^p(r) = \delta\left(rW_{\mathcal{V}}\right)_{o^*} \\
    C_{o^*}^v(r) = \frac{ \left(rW_{\mathcal{V}}\right)_{o^*}}{
    \max\limits_{l=1}^L\max\left(        |(a_N^lW_{\mathcal{V}})_{o^*}|, |(m_N^lW_{\mathcal{V}})_{o^*}|        \right)}
\end{gather}
The definition takes into account both the mapped probability $C_{o^*}^p(r)$ and the normalized logit value $C_{o^*}^v(r)$. 
The rationale for this definition is that, a large mapping probability does not necessarily lead to a significant contribution to the final prediction when the mapping logit value is small. 

The bar chart in Figure \ref{img_attribution} illustrates the average results for LLaVA-V1.5 (7B) \cite{llava} on the E-VQA dataset \cite{MMEdit}, showing that the outputs from deeper layers have a more substantial impact on the key token than those from shallower layers.

\subsection{Visual Representation Attribution}
Noting that the contribution trend of the attention and MLP modules across layers are similar, 
% and that attention outputs depend more directly on the visual representations from the immediate preceding layer compared to MLP outputs, 
we primarily focus on the contributions of visual representations to the next layer attention output. 
Given an attention output $a_N^l=\text{Attn}_l(h_1^{l-1},...,h_{N_v}^{l-1}, ..., h_N^{l-1})$, we perturb each visual representation by set:
\begin{gather}
    \tilde{h}_n^{l-1} = h_n^{l-1} + \epsilon, n\in\{1,...,N_v\}
\end{gather}
where $\epsilon\in \mathcal{R}^{d_h}$ follows $\mathcal{N}\left(0; (3\sigma)^2\right)$ \cite{ROME} and $\sigma$ is the standard deviation of elements in $h_{1:N_v}^{l-1}\in \mathcal{R}^{N_v\times d_h}$. 
Building on the insight that perturbing a significant module input will noticeably alter its output \cite{DBLP:conf/kdd/Ribeiro0G16, DBLP:conf/iccv/FongV17}, we quantify the contribution of each visual representation $h_n^{l-1}$ to the attention output as follows:
\begin{gather}
    C_{a_N^l}(h_n^{l-1}) = \frac{1}{2}\left(1 - \frac{\tilde{a}_N^l\cdot a_N^l}{\|\tilde{a}_N^l\|\cdot\|a_N^l\|}\right)\in[0,1]\label{eq_visual_rep_attribut}\\
    \tilde{a}_N^l = \text{Attn}_l(h_1^{l-1},...,h_{n-1}^{l-1},\tilde{h}_{n}^{l-1}, h_{n+1}^{l-1}, ..., h_N^{l-1})
\end{gather}
The heatmaps in Figure \ref{img_attribution} display the visualization results (For more results, please refer to Appendix \ref{sec_more_visual_rep_Attribution}).
It can be observed that the model focuses on areas highly relevant to the prompt at deep layers. 

% Combining with the previous attribution experiment, we make the hypothesis that VLLM initially aggregates the semantics of a given prompt into the last toke representation at shallow layers, and then extract key information from the visual representation at deep layers to generate responses.
Combining the above two attribution experiments, we hypothesize that the VLLM initially aggregates the semantics of a given prompt into the last token representation at shallow layers, and then extracts key information from visual representations at deep layers to generate responses.

\section{The Proposed Editor}
Inspired by the results of attribution analysis, we devise a VLLM editor named~\emph{VisEdit}. In this section, we first introduce the preliminaries of VLLM editing \cite{MMEdit}. Then, we elaborate on the basic structure of~\emph{VisEdit}. 
Specifically, we introduce an Influence Mapper to help the editor focus on key visual regions based on the prompt, thereby reducing the negative impact on irrelevant visual representations and enhancing editing efficacy. 
Finally, we describe the training process of~\emph{VisEdit}.
\subsection{Preliminaries}
% A VLLM $f_\theta\in\mathcal{F}:\mathcal{X}_v\times\mathcal{X}_t\mapsto \mathcal{O}$ maps an image-prompt pair $(x_v, x_t)$ into text response $o=f_\theta(x_v, x_t)$. 

For a VLLM $f_\theta\in \mathcal{F}$, given an edit sample $(x^e_v, x^e_t, o^e)$ such that $f_\theta(x^e_v, x^e_t)\neq o^e$, an VLLM editor $\text{ME}:\mathcal{F}\times \mathcal{X}_v\times\mathcal{X}_t\times \mathcal{O}\mapsto\mathcal{F}$ outputs an post-edit VLLM $f_{\theta_e}=\text{ME}(f_\theta,x^e_v, x^e_t, o^e)$. A good $\text{ME}$ should meet the following three criteria \cite{MMEdit}:

\textbf{Reliability} assesses the response accuracy of the post-edit model on the edited samples:
\begin{gather*}
% \text{Acc}_{rel}(\mathcal{D}_e) = 
\mathbb{E}_{(x^e_v, x^e_t, o^e)\sim \mathcal{D}_e} \mathbb{I}\left\{f_{\theta_e}(x^e_v, x^e_t) = o^e\right\}
\end{gather*}
where $\mathcal{D}_e$ is edit sample set and $\mathbb{I}$ is the indicator function.

\textbf{Generality} requires the revised model to also make corresponding adjustments to the relevant neighborhoods (e.g., rephrased sentences) of the edited samples, including modal generality and text generality:
\begin{gather*}
\mathbb{E}_{(x^e_v, x^e_t, o^e)\sim \mathcal{D}_e} \mathbb{E}_{x^{mg}_v\sim \mathcal{D}_{mg}(x^e_v)} \mathbb{I}\left\{f_{\theta_e}(x^{mg}_v, x^e_t) = o^e\right\}\\
\mathbb{E}_{(x^e_v, x^e_t, o^e)\sim \mathcal{D}_e} \mathbb{E}_{x^{tg}_t\sim \mathcal{D}_{tg}(x^e_t)} \mathbb{I}\left\{f_{\theta_e}(x^e_v, x^{tg}_t) = o^e\right\}
\end{gather*}
where $\mathcal{D}_{mg}(x_v^e), \mathcal{D}_{tg}(x_t^e)$ respectively represent the neighborhoods of the image $x_v^e$ and prompt $x_t^e$.

\textbf{Locality} requires the revised model to make response consistent with the original model for samples unrelated to the edited samples, including modal locality and text locality:
\begin{equation*}
\begin{aligned}
    \mathbb{E}_{(x^e_v, x^e_t, o^e)\sim \mathcal{D}_e} \mathbb{E}_{(x^{ml}_v, x^{ml}_t, o^{ml})\sim \mathcal{D}_{ml}(x^e_v, x^e_t)} 
    \;\;\;\;\;\;\;\;\;\;\;\;\;\;\;\;\\
    \mathbb{I}\left\{f_{\theta_e}(x^{ml}_v, x^{ml}_t) = f_{\theta}(x^{ml}_v, x^{ml}_t)\right\}
\end{aligned}
\end{equation*}
\begin{equation*}
\begin{aligned}
    \mathbb{E}_{(x^e_v, x^e_t, o^e)\sim \mathcal{D}_e} \mathbb{E}_{(x^{tl}_t, o^{tl})\sim \mathcal{D}_{tl}(x^e_t)} 
    \;\;\;\;\;\;\;\;\;\;\;\;\;\;\;\;\;\;\;\;\;\;\;\;\;\;\;\;\;\\
    \mathbb{I}\left\{f_{\theta_e}(\emptyset, x^{tl}_t) = f_{\theta}(\emptyset, x^{tl}_t)\right\}
\end{aligned}
\end{equation*}
where $\mathcal{D}_{ml}(x^e_v, x^e_t), \mathcal{D}_{tl}(x^e_t)$ respectively represent the multi-modal and text samples irrelevant to the edit sample.

\begin{figure}[t]
\centering
\includegraphics[width=1.\columnwidth]{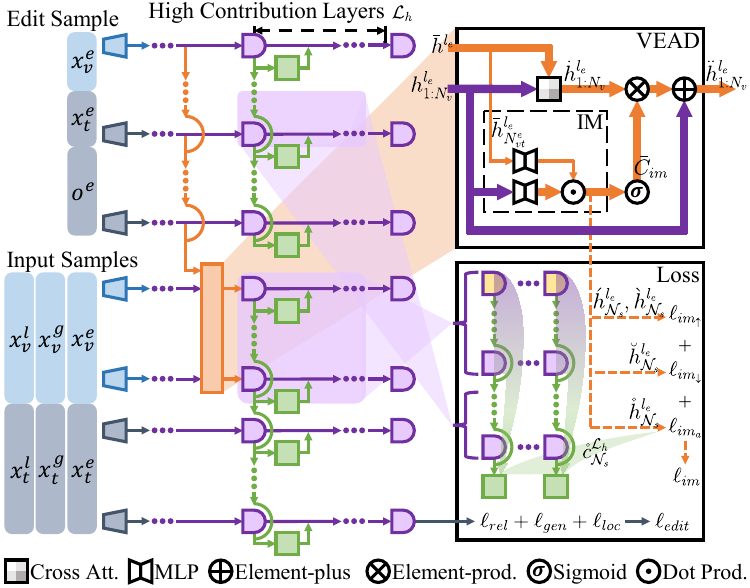} 
\caption{
    Architecture and training loss of~\emph{VisEdit}.
}
\label{img_method}
\vspace{-1em}
\end{figure}
 
\subsection{Architecture of~\emph{VisEdit}}
Based on attribution analysis, we observe that that the VLLM aggregates the semantics of the prompt into the last token in shallow layers; while in deep layers, it extracts information from visual representations related to the prompt to generate responses.
Inspired by this, we devise a VLLM editor, namely \emph{VisEdit}. 
Specifically, as shown in Figure \ref{img_method}, it inserts a Visual Edit ADapter (VEAD) before the high-contribution layers. The adapter modifies the visual representations that the edit sample attends to.
Below, we assume that VEAD is inserted at the $l_e$-th layer of the VLLM and specify two steps in detail: (1) Compute edit signal from an edit sample. (2) Use the edit signal to adapt the model for future inputs.

\noindent\textbf{Compute Edit Signals:}  
Given an edit sample $(x^e_v, x^e_t, o^e)$, VEAD feeds $(x_v^e,  x_t^e \oplus o^e)$ into $f_\theta$ and takes the $l_e$-th layer output $\bar{h}^{l_e}\in\mathcal{R}^{(N_v+N_t^e+N_o^e)\times d_h}$ as the edit signal.
% Thus, the $f_\theta$ equipped with VEAD is edited to $f_{\theta_e}$.
Here, $\oplus$ denotes string concatenation, and $N_v, N_t^e, N_o^e$ are the dimensions of representations of $x^e_v, x^e_t, o^e$ respectively.

\noindent\textbf{Adapt Hidden State with Influence Mapper:}
% After capturing the edit signal $\bar{h}^{l_e}$, given an sample $(x_v, x_t)$ that is input to $f_{\theta_e}$, its $l_e$-th layer visual representations $h^{l_e}_{1:N_v}\in \mathcal{R}^{N_v\times d_h}$ will be modified by VEAD to meet the editing requirements. 
With the edit signal $\bar{h}^{l_e}$ defined, we now demonstrate how to edit the original visual representation $h^{l_e}_{1:N_v}\in \mathcal{R}^{N_v\times d_h}$ for a given $(x_v, x_t)$. 
Specifically, a cross-attention operation is applied to integrate the edit signal into $h^{l_e}_{1:N_v}$, i.e.,
\begin{gather}
\dot{h}^{l_e}_{1:N_v} = \delta\left(h^{l_e}_{1:N_v}W_1\left(\bar{h}^{l_e}W_2\right)^T\right)\bar{h}^{l_e}W_3
\end{gather}
where $W_1\in\mathcal{R}^{d_h\times d_a}, W_2\in\mathcal{R}^{d_h\times d_a}, W_3\in\mathcal{R}^{d_h\times d_h}$ are the projection matrices, with related biases omitted here.
$\delta$ is the softmax function, and $d_a$ is the dimension of modules in VEAD.
% To minimize the interference of the above integration on unrelated inputs or the visual representations with low contribution in related inputs, we incorporate an Influence Mapper (IM) module $f_{im}$ to learn the attention pattern that VLLM extracts visual features from key regions related to the edit prompt, thus to control the scaling of adaptation. 
% To make sure that the edit signal is not applied on the visual region irrelevant with the edit prompt or the locality samples, we further incorporate an Influence Mapper (IM) module $f_{im}$ to control the intensity of the adaptation. 
% It enhances the efficacy of editing and reduces negative impact to irrelevant visual representations. 
To make sure that the edit signal is not applied to locality samples or visual regions irrelevant to the edit prompt within relevant samples, we further incorporate an Influence Mapper (IM) module $f_{im}$ to control the edit intensity of the adaptation.
Based on the previous attribution analysis, we design IM to use the last token of the edit prompt to generate the edit intensity $\bar{C}_{im}\in\mathcal{R}^{N_v\times 1}$ for the visual regions, defined as follows:
\begin{equation}
\begin{aligned}
    \bar{C}_{im} &= \sigma\left(f_{im}\left(h^{l_e}_{1:N_v}, \bar{h}^{l_e}_{N_{vt}^e}\right)\right) \\
           &=\sigma\left( f_{\mu_1}\left(h^{l_e}_{1:N_v}\right)\cdot f_{\mu_2}\left(\bar{h}^{l_e}_{N_{vt}^e}\right)^\top\right)
\end{aligned}
\end{equation}
where $N_{vt}^e = N_v+N_t^e$, and thus $\bar{h}^{l_e}_{N_{vt}^e}\in\mathcal{R}^{d_h}$ is the representation according to the last token embedding of $x_t^e$ in $\bar{h}^{l_e}$. 
$f_{\mu_1}, f_{\mu_2}$ are two-layers MLPs, mapping dimension to $d_a$. $\sigma$ is the sigmoid function.
Finally, the adapted visual representations $\ddot{h}^{l_e}_{1:N_v}$ is formulated as:
\begin{gather}
 \ddot{h}^{l_e}_{1:N_v} = h^{l_e}_{1:N_v} + \dot{h}^{l_e}_{1:N_v} \times \bar{C}_{im}
    \label{eq_VEAD_output}
\end{gather}
where $\times$ indicates element-wise product with broadcast.

\subsection{Training Process of~\emph{VisEdit}}
The training loss of~\emph{VisEdit} includes the editing loss and the IM loss, designed to meet the editing requirements and enable IM to learn the edit intensity, respectively.

\noindent\textbf{Editing Loss:} 
Given an edit sample $(x_v^e,x_t^e,o^e)$, corresponding generality samples $x^{mg}_v, x^{tg}_t$, and locality samples $(x^{ml}_v, x^{ml}_t, o^{ml}),(x^{tl}_t, o^{tl})$, 
the editing loss is the sum of the reliability loss $\ell_{rel}$, the generality loss $\ell_{gen}$, and the locality loss $\ell_{loc}$, defined as follows:
\begin{equation}
    \ell_{edit} = \ell_{rel} + \ell_{gen} + \ell_{loc}
\end{equation}
where
\begin{gather}
    \ell_{rel} = -\log f_{\theta_e}\left(o^e|x_v^e,x_t^e\right)
\end{gather}
\begin{equation}
\begin{aligned}
    \ell_{gen} \!=\! -\log f_{\theta_e}\left(o^e|x_v^{mg},x_t^e\right)
                      -\log f_{\theta_e}\left(o^e|x_v^e,x_t^{tg}\right)
    % \ell_{gen} = -\log\left(f_{\theta_e}\left(o^e|x_v^{mg},x_t^e\right)\right)\\ 
    % % \;\;\;\;\;\;\;\;\;\;\;\;\;\;\;\;\;\;\;\;\;\;\;\;\;\;\;\;\;\;\\
    %                   -\log\left(f_{\theta_e}\left(o^e|x_v^e,x_t^{tg}\right)\right)
\end{aligned}
\end{equation}
\begin{equation}
\begin{aligned}
    \ell_{loc} \!= &
    \text{KL}\left(
            f_{\theta}\left(o^{ml}|x_v^{ml},x_t^{ml}\right) ||
            f_{\theta_e}\left(o^{ml}|x_v^{ml},x_t^{ml}\right)
        \right)
    \\&+\text{KL}\left(
            f_{\theta}\left(o^{tl}|\emptyset,x_t^{tl}\right) ||
            f_{\theta_e}\left(o^{tl}|\emptyset,x_t^{tl}\right)
        \right)\\
\end{aligned}
\end{equation}
Here $\text{KL}$ denotes the Kullback-Leibler divergence.

\begin{table*}[t!]
\centering
    % \scriptsize
    % \setlength{\tabcolsep}{5.9pt}
    % \renewcommand{\arraystretch}{0.45}
    \footnotesize
    \setlength{\tabcolsep}{2.8pt}
    \renewcommand{\arraystretch}{0.55}

\begin{tabular}{cccccccccccccc}
\toprule
 &  & \multicolumn{6}{c}{\textbf{E-VQA}} & \multicolumn{6}{c}{\textbf{E-IC}} \\
\multirow{-3}{*}{\textbf{Backbone}} & \multirow{-3}{*}{\textbf{Editor}} & \textbf{Rel.} & \textbf{T-Gen.} & \textbf{M-Gen.} & \textbf{T-Loc.} & \textbf{M-Loc.} & \textbf{Average} & \textbf{Rel.} & \textbf{T-Gen.} & \textbf{M-Gen.} & \textbf{T-Loc.} & \textbf{M-Loc.} & \textbf{Average} \\
\midrule
 & FT-V & 24.01 & 16.00 & 20.22 & 100.00 & 88.65 & 49.78$_{(\pm0.47)}$ & 42.11 & 40.74 & 36.43 & 100.00 & 89.73 & 61.80$_{(\pm0.40)}$ \\
 & FT-L & 24.86 & 16.39 & 20.57 & 98.92 & 89.61 & 50.07$_{(\pm0.60)}$ & 41.61 & 40.31 & 37.41 & 99.35 & 88.70 & 61.48$_{(\pm0.77)}$ \\
 & \cellcolor[HTML]{E7E6E6}KE & \cellcolor[HTML]{E7E6E6}67.80 & \cellcolor[HTML]{E7E6E6}63.00 & \cellcolor[HTML]{E7E6E6}66.17 & \cellcolor[HTML]{E7E6E6}97.32 & \cellcolor[HTML]{E7E6E6}45.89 & \cellcolor[HTML]{E7E6E6}68.04$_{(\pm0.00)}$ & \cellcolor[HTML]{E7E6E6}69.00 & \cellcolor[HTML]{E7E6E6}62.80 & \cellcolor[HTML]{E7E6E6}61.22 & \cellcolor[HTML]{E7E6E6}96.21 & \cellcolor[HTML]{E7E6E6}45.55 & \cellcolor[HTML]{E7E6E6}66.96$_{(\pm0.00)}$ \\
 & \cellcolor[HTML]{E7E6E6}IKE & \cellcolor[HTML]{E7E6E6}\textbf{99.95} & \cellcolor[HTML]{E7E6E6}91.59 & \cellcolor[HTML]{E7E6E6}92.33 & \cellcolor[HTML]{E7E6E6}13.16 & \cellcolor[HTML]{E7E6E6}1.88 & \cellcolor[HTML]{E7E6E6}59.78$_{(\pm0.00)}$ & \cellcolor[HTML]{E7E6E6}96.70 & \cellcolor[HTML]{E7E6E6}78.20 & \cellcolor[HTML]{E7E6E6}83.15 & \cellcolor[HTML]{E7E6E6}13.36 & \cellcolor[HTML]{E7E6E6}2.17 & \cellcolor[HTML]{E7E6E6}54.72$_{(\pm0.00)}$ \\
 & \cellcolor[HTML]{E7E6E6}SERAC & \cellcolor[HTML]{E7E6E6}91.20 & \cellcolor[HTML]{E7E6E6}91.40 & \cellcolor[HTML]{E7E6E6}89.81 & \cellcolor[HTML]{E7E6E6}100.00 & \cellcolor[HTML]{E7E6E6}0.33 & \cellcolor[HTML]{E7E6E6}74.55$_{(\pm0.00)}$ & \cellcolor[HTML]{E7E6E6}94.40 & \cellcolor[HTML]{E7E6E6}96.00 & \cellcolor[HTML]{E7E6E6}91.49 & \cellcolor[HTML]{E7E6E6}100.00 & \cellcolor[HTML]{E7E6E6}0.47 & \cellcolor[HTML]{E7E6E6}76.47$_{(\pm0.00)}$ \\
 & \cellcolor[HTML]{E7E6E6}MEND & \cellcolor[HTML]{E7E6E6}92.60 & \cellcolor[HTML]{E7E6E6}90.80 & \cellcolor[HTML]{E7E6E6}91.94 & \cellcolor[HTML]{E7E6E6}96.07 & \cellcolor[HTML]{E7E6E6}65.15 & \cellcolor[HTML]{E7E6E6}87.31$_{(\pm0.00)}$ & \cellcolor[HTML]{E7E6E6}65.00 & \cellcolor[HTML]{E7E6E6}38.00 & \cellcolor[HTML]{E7E6E6}36.19 & \cellcolor[HTML]{E7E6E6}92.67 & \cellcolor[HTML]{E7E6E6}55.72 & \cellcolor[HTML]{E7E6E6}57.52$_{(\pm0.00)}$ \\
 & TP & 68.31 & 60.88 & 56.35 & 98.49 & 85.27 & 73.86$_{(\pm0.99)}$ & 49.71 & 49.03 & 45.46 & 93.88 & 80.88 & 63.79$_{(\pm1.11)}$ \\
 & LTE & 97.74 & 97.21 & 96.35 & 94.34 & 84.99 & 94.13$_{(\pm0.97)}$ & 96.69 & 95.26 & 94.06 & 95.25 & 87.68 & 93.79$_{(\pm1.05)}$ \\
\multirow{-14.5}{*}{\begin{tabular}[c]{@{}c@{}}BLIP2-OPT\\      (2.7B)\end{tabular}} & \textit{VisEdit} & 98.83 & \textbf{98.63} & \textbf{97.90} & \textbf{100.00} & \textbf{92.30} & \textbf{97.53}$_{(\pm0.70)}$ & \textbf{97.06} & \textbf{96.83} & \textbf{94.85} & \textbf{100.00} & \textbf{91.74} & \textbf{96.10}$_{(\pm0.94)}$ \\
\midrule
 & FT-V & 31.68 & 29.96 & 26.68 & 100.00 & 91.23 & 55.91$_{(\pm0.71)}$ & 52.85 & 51.57 & 48.63 & 100.00 & 92.55 & 69.12$_{(\pm0.29)}$ \\
 & FT-L & 31.78 & 30.02 & 26.91 & 99.94 & 92.03 & 56.14$_{(\pm2.13)}$ & 53.00 & 51.02 & 49.29 & 98.91 & 94.89 & 69.42$_{(\pm1.71)}$ \\
 & KE & 85.86 & 84.00 & 82.23 & 93.57 & 73.06 & 83.74$_{(\pm1.25)}$ & 83.54 & 82.15 & 81.12 & 92.46 & 73.83 & 82.62$_{(\pm0.88)}$ \\
 & IKE & 91.35 & 90.84 & 91.08 & 60.18 & 51.08 & 76.91$_{(\pm1.42)}$ & 93.72 & 88.37 & 76.99 & 76.60 & 64.90 & 80.12$_{(\pm1.18)}$ \\
 & SERAC & 82.51 & 81.60 & 80.05 & 100.00 & 57.48 & 80.33$_{(\pm1.58)}$ & 43.08 & 42.37 & 42.85 & 100.00 & 7.63 & 47.19$_{(\pm0.83)}$ \\
 & MEND & 92.30 & 92.16 & 92.10 & 90.30 & 81.13 & 89.60$_{(\pm2.36)}$ & 93.76 & 93.46 & 92.14 & 91.60 & 87.59 & 91.71$_{(\pm1.42)}$ \\
 & TP & 38.68 & 36.27 & 31.26 & 95.31 & 91.41 & 58.59$_{(\pm1.32)}$ & 59.07 & 57.01 & 55.51 & 64.79 & 89.26 & 65.13$_{(\pm1.85)}$ \\
 & LTE & 94.16 & 93.54 & 93.06 & 83.76 & 81.65 & 89.23$_{(\pm1.90)}$ & 93.60 & 92.38 & 91.18 & 85.54 & 88.49 & 90.24$_{(\pm1.90)}$ \\
\multirow{-14.5}{*}{\begin{tabular}[c]{@{}c@{}}LLaVA-V1.5\\      (7B)\end{tabular}} & \textit{VisEdit} & \textbf{95.99} & \textbf{95.78} & \textbf{94.71} & \textbf{100.00} & \textbf{94.12} & \textbf{96.12}$_{(\pm0.97)}$ & \textbf{95.27} & \textbf{94.64} & \textbf{93.57} & \textbf{100.00} & \textbf{96.20} & \textbf{95.94}$_{(\pm0.90)}$\\
\bottomrule
\end{tabular}

\caption{Editing performance of BLIP2-OPT and LLaVA-V1.5 evaluated on E-VQA and E-IC datasets. ``Rel.'', ``T/M-Gen.'' and ``T/M-Loc.'' stand for reliability, text/modal generality, and text/modal locality, respectively. Results with a gray background are taken from \citet{MMEdit}. The t-tests demonstrate our improvements are statistically significant with $p < 0.05$ level.}
% \vspace{-0.5em}
\label{tab_main_exp}
\vspace{-1em}
\end{table*}

\noindent\textbf{IM Loss:} 
As shown in Eq.~\ref{eq_VEAD_output}, IM controls the edit intensity. For reliability and generality samples, the intensity should be large, whereas for locality samples, it should be small. Therefore, in each training iteration, we randomly sample a portion of visual representations to compute the corresponding loss:
\begin{equation}
\begin{aligned}
\ell_{im_{\uparrow}} = \frac{1}{|\mathcal{N}_s|}\sum\limits_{n\in\mathcal{N}_s}&\left(-\log \sigma\left( f_{im}\left(\acute{h}^{l_e}_{n}, \bar{h}^{l_e}_{N_{vt}^e}\right)\right)\right.\\
            &\left.\;\,-\log \sigma\left(f_{im}\left(\grave{h}^{l_e}_{n}, \bar{h}^{l_e}_{N_{vt}^e}\right)\right)
            \right)
\end{aligned}
\end{equation}
\begin{equation}
\begin{aligned}
\ell_{im_{\downarrow}} =  \frac{1}{|\mathcal{N}_s|}\sum\limits_{n\in\mathcal{N}_s} \!\!-\! \log \left(1- \sigma\left(f_{im}\left(\breve{h}^{l_e}_{n}, \bar{h}^{l_e}_{N_{vt}^e}\right)\right)\right)
\end{aligned}
\end{equation} 
where $\mathcal{N}_s\subset\{1, ..., N_v\}$ is the sample index, and $\acute{h}, \grave{h},\breve{h}$ are the representations of reliability, generality, and locality samples, respectively.
% To guide IM to put attention on visual regions highly relevant to the edit prompt, we regularize IM to approximate the attribution results that combine sampled visual representations with edit prompt representations.
To guide IM to put attention on visual regions highly relevant to the edit prompt, we regularize IM to approximate the VLLM-focused visual regions, through visual representation attribution on the sampled visual representations based on the edit prompt.
Assuming the set of high-contribution layers is $\mathcal{L}_{h}\subset \{1,...,L\}$, we first replace the visual representations of the edit sample, i.e., $\bar{h}$, from layers in $\mathcal{L}_{h}$ with the counterpart of $\mathring{h}$ randomly selected from $\{\acute{h},\grave{h}\}$. 
Then, following Equation \ref{eq_visual_rep_attribut}, we compute the contribution of visual representations in $\mathring{h}$ to the edit prompt as:
\begin{gather}
    \mathring{c}_n^l = C_{\bar{a}^{l}_{N_{vt}^e}}\left(\mathring{h}^{l}_{n}\right)\in[0,1], l\in \mathcal{L}_{h}, n\in\mathcal{N}_s
\end{gather}
where $\bar{a}_{N_{vt}^e}^l$ represents the $l$-th layer attention output at the last token of the edit prompt.
% Similarly, we obtain the attribution results for generality sample $\grave{c}_n^l \in[0,1]$.
Then, we minimize the discrepancy between IM’s mapping results and the attribution results averaged across layers using cross-entropy loss: 
\begin{gather}
    % \ell_{im_{a}} = \psi(\acute{c}, \acute{h}) + \psi(\grave{c}, \grave{h})\\
    \ell_{im_{a}} \!=\! \sum_{\substack{n \in \mathcal{N}_s \\ l \in \mathcal{L}_h}} \frac{-\mathring{c}^l_n}{|\mathcal{L}_{h}|\sum\limits_{j\in\mathcal{N}_s}\mathring{c}^l_j}\log \delta \left(f_{im}\left(\mathring{h}^{l_e}_{\mathcal{N}_s},\bar{h}^{l_e}_{N_{vt}^e}\right)\right)_n%\notag
\end{gather} 
% \begin{gather}
%     \ell_{im_{a}} = \psi(\acute{c}, \acute{h}) + \psi(\grave{c}, \grave{h})\\
%     \psi(\alpha, \beta) = \sum_{\substack{n \in \mathcal{N}_s \\ l \in \mathcal{L}_h}} \frac{-\alpha^l_n}{|\mathcal{L}_{h}|\sum\limits_{j\in\mathcal{N}_s}\alpha^l_j}\log \delta \left(f_{im}\left(\beta^{l_e}_{\mathcal{N}_s},\bar{h}^{l_e}_{N_{vt}^e}\right)\right)_n\notag
% \end{gather} 
where $\mathring{h}^{l_e}_{\mathcal{N}_s} \in\mathcal{R}^{|\mathcal{N}_s|\times d_h}$ indicates the sampled representations from $l_e$-th layer, and $\delta(\cdot)_n$ represents the $n$-th element of the vector outputted from softmax.
% \grave{c}^l\in[0,1]^{|\mathcal{N}_s|\times 1}$ and $\acute{h}^{l_e}_{\mathcal{N}_s},\grave{h}^{l_e}_{\mathcal{N}_s}\in\mathcal{R}^{|\mathcal{N}_s|\times d_h}$. 
Thus, the IM loss is formulated as $\ell_{im} = \ell_{im_\uparrow} + \ell_{im_\downarrow} + \ell_{im_a}$. 

The total loss of~\emph{VisEdit} is: $\ell_{total} = \ell_{edit} + \ell_{im}$. 
% where $\lambda$ is the coefficient to balance the two losses. 
During training, we freeze the parameters of the VLLM and only the parameters in VEAD are updated.

\section{Experiments}
% In this section, we first evaluate the editing performance of~\emph{VisEdit}. Subsequently, we conduct exploratory and ablation experiments on modules. 
\subsection{Experimental Settings}

\noindent\textbf{Datasets:}
Following \citet{MMEdit}, we employ E-VQA (Editing Visual Question Answering) and E-IC (Editing Image Caption) as evaluation datasets.

\noindent\textbf{VLLM Backbones:} To ensure a comprehensive evaluation, we consider both the parameter size and model architecture when selecting VLLM backbones for editing, including BLIP2-OPT (2.7B) \cite{BLIP2}, LLaVA-V1.5 (7B) \cite{llava}, and MiniGPT-4 (7B) \cite{MiniGPT-4}.

\noindent\textbf{Baseline Editors:} To the best of our knowledge, there are currently no editors specifically designed for VLLMs. 
% Therefore, editors developed for LLMs are adapted to the VLLM scenario, 
Therefore, most existing works \cite{MMEdit, DBLP:journals/corr/abs-2402-14835} applied LLM editors in the VLLM case, including FT-V (Fine-tunes visual encoder), TF-L (Fine-tunes last layer of the language model), KE \cite{KnowledgeEditor}, IKE \cite{InContextKnowledgeEdit}, SERAC \cite{SERAC}, MEND \cite{MEND}, TP \cite{T-Patcher}, and LTE \cite{LTE}. 
For experimental setup details, as well as the model hyperparameters and training specifics, please refer to Appendix \ref{Appendix_detail_exp_settings}.

Using the above experimental settings, we comprehensively evaluate the editing performance and conduct in-depth quantitative analyses of VEAD internals to validate that it effectively incorporates insights from the attribution analysis.

\subsection{Analysis of Editing Performance}
The overall editing performance is exhibited in Table \ref{tab_main_exp} and Appendix \ref{sec_minigpt4_editing_performance}. Here we analyze Table \ref{tab_main_exp} from different perspectives.

\noindent\textbf{From the perspective of editors}, \emph{VisEdit} demonstrates the best performance.  LTE also excels because it fine-tunes the entire LLM.
However, unlike pure language models, the combination of visual and language representations makes it harder for the model to learn the mode following editing instructions.

\noindent\textbf{From the perspective of backbones}, VEAD performs slightly better on BLIP2. We ascribe this to VEAD intervening only in a local part of the VLLM's data pathway, where the effect of this intervention would be relatively more significant in a smaller model.

\noindent\textbf{From the perspective of datasets}, most editors perform better on E-VQA than on E-IC. This is due to E-VQA involving the correction of a few key tokens, while E-IC requires editing captions that convey complete image information, making it more challenging.

\noindent\textbf{From the perspective of metrics}, editors with lower average scores often show that high reliability and generality cannot coexist with locality. 
Additionally, because FT-V and \emph{VisEdit} only edit the visual component, and SERAC uses classifiers to distinguish pure text input, they effectively avoid interference from the editing process on the text locality samples.
% Additionally, in terms of text locality, since FT-V and our \emph{VisEdit} only edit the visual part, they do not interfere with the input scenario of pure language. SERAC, by training a classifier, can effectively distinguish between edited samples containing visual representations and text locality inputs without visual representations.

\begin{table}[t!]
    \footnotesize
    \setlength{\tabcolsep}{3.85pt}
    \renewcommand{\arraystretch}{0.51}

\begin{tabular}{lcccccc}
\toprule
\textbf{Layer} & \textbf{Rel.}  & \textbf{T-Gen.} & \textbf{M-Gen.} & \textbf{T-Loc.} & \textbf{M-Loc.} & \textbf{Average} \\
\midrule
0                  & 95.10          & 93.70           & 94.43           & 100.00          & 84.51           & 93.55            \\
5                  & 95.16          & 94.63           & 94.49           & 100.00          & 86.51           & 94.16            \\
10                 & 96.36          & 95.79           & 95.60           & 100.00          & 85.83           & 94.72            \\
15                 & 97.69          & 97.48           & 97.08           & 100.00          & 92.20           & 96.89            \\
19$^*$                 & \textbf{98.83} & \textbf{98.63}  & \textbf{97.90}  & \textbf{100.00} & \textbf{92.30}  & \textbf{97.53}   \\
25                 & 97.54          & 96.97           & 95.87           & 100.00          & 88.63           & 95.80            \\
30                 & 86.22          & 84.18           & 83.62           & 100.00          & 85.98           & 88.00           \\
\bottomrule
\end{tabular}

\caption{Editing results on E-VQA when VEAD is attached to different layers of BLIP2-OPT (2.7B), where the layer marked with an asterisk represents our selection based on the first attribution experiment.}
\label{tab_vead_layer_exploration}
\vspace{-1em}
\end{table}

\begin{figure}[t]
\centering
\includegraphics[width=1.\columnwidth]{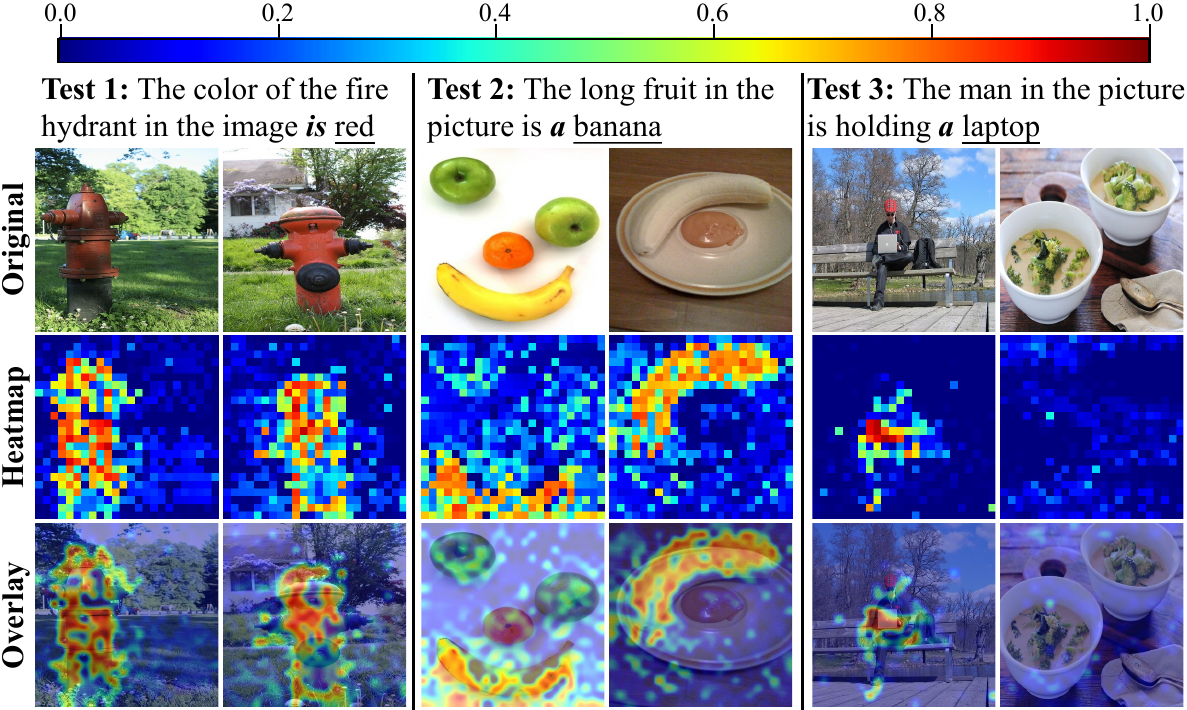} 
\caption{The visualization of IM module in VEAD integrated into LLaVA-V1.5. In each test, VEAD is first edited using the left image along with the prompt. Then, the outputs of IM are visualized for both images.}
\label{img_IM_Visualization}
\vspace{-1em}
\end{figure}

\begin{figure}[t]
\centering
\includegraphics[width=1.\columnwidth]{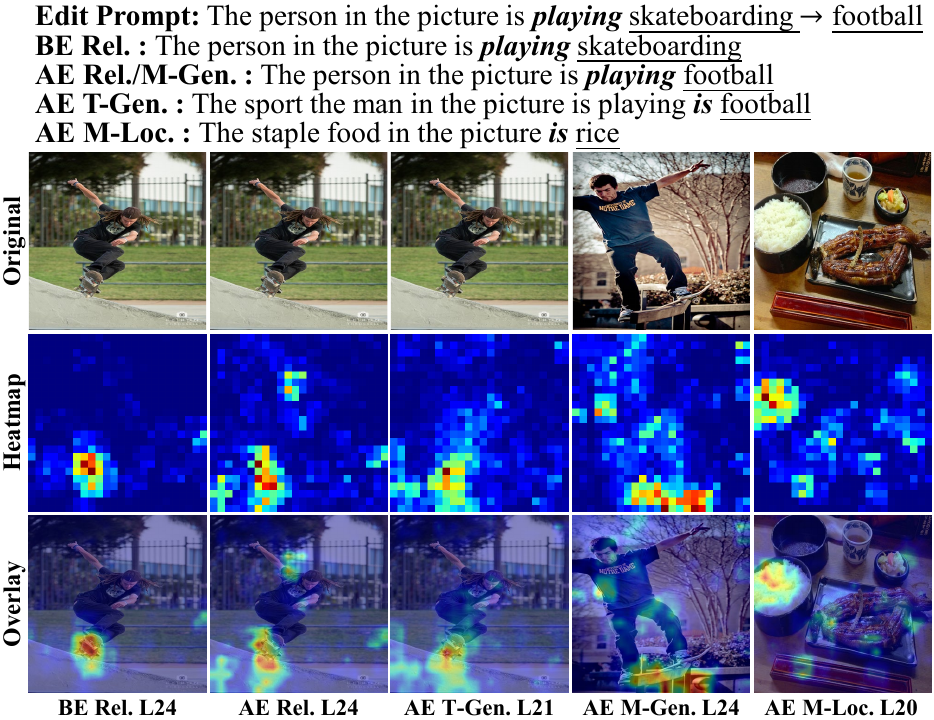} 
\caption{
% Post-\emph{VisEdit}-edited visual representation attribution visualization on a counterfactual edit sample. 
Visualization of visual representation attribution after \emph{VisEdit} edits a counterfactual sample. ``\textbf{BE}'' and ``\textbf{AE}'' indicate Before Editing and After Editing respectively. \textbf{L*} indicates the layer index. }
\label{img_vead_post_edit_attribution_visualization}
\vspace{-1em}
\end{figure}

\subsection{Analysis of VEAD Internals}
We conduct three sets of experiments to demonstrate that VEAD fully incorporates insights from the attribution analysis. First, we compare VEAD's editing performances across different layers. Next, we visualize the outputs of IM to demonstrate that it attends to the visual regions important for the edit prompt. Finally, we verify whether VEAD successfully adapts the visual representations critical for response generation.

\noindent\textbf{Editing Performance in Different Layers:} 
% First, we explore the determination of the layer attached by VEAD, based on the first attribution experiment.
Table \ref{tab_vead_layer_exploration} displays the editing performance when VEAD is applied to various layers of BLIP2.
We can observe that the editing performance first gets better as we go deeper until layer 19, and then gets worse for later layers. 
We hypothesize that the edit signal in too shallow layers is prone to lose before reaching the deep layers that highly contribute to prediction, while for cases in too deep layers, the signal cannot have a substantial enough impact on the final prediction.
% It can be observed that the editing performance gradually improves as the layer index increases, peaking at layer 19.
% We hypothesize that one reason is the adaptations of model responses struggle to benefit from modifications of visual representations occurring in shallow layers with low contributions, and another is that the editing information is prone to loss within the data pathways of these layers, thereby reducing the editing gains in high-contribution layers.
% Subsequently, the further increase of the layer index will skip some high-contribution layers, preventing them from capturing editing information and resulting in decreased editing performance.

\noindent\textbf{Visualization of IM Module:} 
% The IM module is integrated into VEAD based on the second attribution experiment. To verify whether IM can learn the visual attention patterns of the VLLM, secondly, 
We visualize the outputs of IM in Figure \ref{img_IM_Visualization} (For more results and analysis, please refer to Appendix \ref{sec_additional_im_visualization}). 
Specifically, Test 1 and Test 2 indicate that IM indeed picks the relevant visual region utilizing the overall semantics of the edit prompt, even when the visual objects in the edited samples are not entirely consistent with those in the input samples, e.g., the peeled and unpeeled bananas in test 2.
% Test 3 demonstrates that visual representations in images unrelated to the prompt are given low attention by VLLM. 
Test 3 demonstrates that the visual areas of samples unrelated to the edit sample will receive low edit intensities from IM, because of $\ell_{im_a}$ and $\ell_{im_\downarrow}$.

\noindent\textbf{Instance Analysis for VEAD:} 
% The above experiments show that IM learns the visual attention patterns of VLLM. Furthermore, we validate that VEAD edits the visual representations at the attended region.
Figure \ref{img_vead_post_edit_attribution_visualization} illustrates a counterfactual edit example where VEAD forces LLaVA-V1.5 to follow the knowledge even if it is incorrect (For more results and analysis, please refer to Appendix \ref{sec_additional_instance_analysis_VEAD}).
% By comparing the visualization results of the reliability samples before editing with those of the reliability and generality samples after editing, it shows the editing process does not significantly change the visual regions VLLM focuses on for similar prompts, even if the ``skateboarding'' is counterfactually edited to ``football''.
We first compare the visualization results of three cases: 
reliability samples before editing (BE Rel.),  reliability samples after editing (AE Rel.), and generality samples after editing (AE T/M-Gen.).
We observe that the editing process does not significantly change the visual regions VLLM focuses on for similar prompts, even if the ``skateboarding'' is counterfactually edited to ``football''.
% The above observation indicates that VEAD indeed alters the visual representations in response to the prompt to those that highly contribute to predicting the edited token (e.g., the ``football''), thereby validating the design objective of our model.
The above observation indicates that VEAD adapts the visual representations in the skateboard region to new visual representations for football, and furthermore alters the final prediction to football, thereby validating the design objective of our model.
Additionally, the visualization of modal locality samples shows that VEAD maintains VLLM's original attention to visual objects unrelated to the edit.

\begin{table}[t!]
    % \small
    % \setlength{\tabcolsep}{3.4pt}
    % \renewcommand{\arraystretch}{0.9}
    
    \footnotesize
    \setlength{\tabcolsep}{3.5pt}
    \renewcommand{\arraystretch}{0.51}

\begin{tabular}{lcccccc}
\toprule
\textbf{Settings} & \textbf{Rel.} & \textbf{T-Gen.} & \textbf{M-Gen.} & \textbf{T-Loc.} & \textbf{M-Loc.} & \textbf{Average} \\
\midrule
VEAD & \textbf{95.99} & \textbf{95.78} & \textbf{94.71} & \textbf{100.00} & 94.12 & \textbf{96.12} \\
- $\ell_{im_\downarrow}$ & 95.50 & 94.38 & 93.87 & 100.00 & 86.72 & 94.09 \\
- $\ell_{im_\uparrow}$ & 94.13 & 93.82 & 93.58 & 100.00 & 93.71 & 95.05 \\
- $\ell_{im_a}$ & 93.70 & 93.05 & 92.51 & 100.00 & 90.18 & 93.89 \\
-  IM & 93.13 & 92.58 & 91.89 & 100.00 & 83.84 & 92.29 \\
-  CA & 33.79 & 31.98 & 28.94 & 100.00 & \textbf{98.98} & 58.74\\
\bottomrule
\end{tabular}
\caption{Ablation study of VEAD.}
\label{tab_ablation}
\vspace{-1em}
\end{table}

\subsection{Ablation Study}
Table \ref{tab_ablation} shows the results of the ablation study for VEAD editing LLaVA-V1.5 on E-VQA. Overall, the removal of losses resulted in some degradation in reliability, generality, and locality. 
Specifically, removing $\ell_{im_\downarrow}$ caused significant damage to modal locality, as it makes IM output a smaller edit intensity for samples unrelated to the edit sample. 
$\ell_{im_\uparrow}$ and $\ell_{im_a}$ encourage IM to apply higher edit intensities to visual objects that are consistent with the semantics of the edit prompt; their removal leads to a relatively significant degradation in both reliability and generality scores.
Additionally, removing the entire IM module fixes the edit intensity to 1, preventing VEAD from focusing on specific visual regions for modifications and increasing erroneous interventions in unrelated samples. Since the CA (Cross-Attention) module is the key component for integrating the edit signal, its removal essentially caused VEAD to lose its editing capability. These results show the effectiveness of the modules and loss functions designed in VEAD.

\section{Conclusion}
% This paper proposes a two-step attribution method based on contribution allocation and noise perturbation to measure the impact of visual representations on token prediction in VLLM. The attribution results show that the mid-to-late layers of VLLM pay particular attention to visual regions highly relevant to the prompt semantics. Based on the attribution results, we propose a novel VLLM editor, \emph{VisEdit}. A visual editing adapter is inserted before the high-contribution layers to integrate editing information into the visual representations. Inside the editor, a novel IM module is introduced to enhance the edit signal in the most relevant visual regions and ignore the irrelevant ones. In VLLM editing experiments, \emph{VisEdit} demonstrates exceptional editing performance. For future work, we plan to conduct further research on the attribution results of VLLM and explore expanding VEAD to multiple editing scenarios, such as by incorporating retrieval mechanisms.

This paper proposes a two-step attribution method combining contribution allocation and noise perturbation to measure the impact of visual representations on token prediction in VLLM. Results reveal that mid-to-late layers of VLLM focus on visual regions closely related to prompt semantics. Leveraging these insights, we introduce \emph{VisEdit}, a novel VLLM editor with a visual editing adapter that integrates editing information into high-contribution layers. The editor incorporates an IM module to enhance relevant edit signals while ignoring irrelevant ones. \emph{VisEdit} achieves outstanding performance in VLLM editing tasks. Future work will explore deeper insights into VLLM attributions and extend VEAD to broader editing scenarios.

% \clearpage

\section*{Acknowledgments}
This work is supported by the National Science and Technology Major Project (Grant No. 2022ZD0120302).

\bibliography{aaai25}

\clearpage
\appendix

\section{Notations}
\label{appendix_notations}
The symbols and notations for variables and functions are presented in Table \ref{tab_notations_variable} and Table \ref{tab_notations_function}, respectively.

\begin{table*}[t!]
    \setlength{\tabcolsep}{3.5pt}
    \renewcommand{\arraystretch}{0.51}

\begin{tabular}{ccl}
\toprule
\textbf{Variables} & \textbf{Domain} & \textbf{Explanation \& Variants} \\
\midrule
$(x_v, x_t)$ & $(\mathcal{X}_v, \mathcal{X}_t)$ & \begin{tabular}[c]{@{}l@{}}an input image-text pair to a VLLM. \\ 
$x^e_\cdot$ is the input for editing.\\
$x^{mg}_\cdot,x^{tg}_\cdot$ are the inputs for modal/text generality, respectively.\\
$x^{ml}_\cdot,x^{tl}_\cdot$ are the inputs for modal/text locality, respectively. \end{tabular} \\
\midrule
$o$ & $\mathcal{O}$ & \begin{tabular}[c]{@{}l@{}}a text response from a VLLM. \\ $o^*$ is the key token to be predicted in attribution experiments. \\ $o^e,o^{ml},o^{tl}$ are the predicting objects for editing and modal/text locality samples, respectively.\end{tabular} \\
\midrule
$\boldsymbol{\varepsilon}$ & $\mathcal{R}^{N\times d_h}$ & \begin{tabular}[c]{@{}l@{}}embedding of an input image-text pair. \\ $\boldsymbol{\varepsilon}_v, \boldsymbol{\varepsilon}_t$ are the embeddings for vision and text input, respectively.\end{tabular} \\
\midrule
$y$ & $\mathcal{R}^{\|\mathcal{V}\|}$ & predicted probability distribution. \\
\midrule
$N$ & $\mathcal{Z}^+$ & \begin{tabular}[c]{@{}l@{}}length of embeddings sequence. \\ $N^\cdot_v,N^\cdot_t,N^\cdot_{vt}$ are respectively the lengths of the vision, the text, and their concatenated hidden states. \\ $N^e_\cdot$ is the representation length of an editing sample.\end{tabular} \\
\midrule
$L$ & $\mathcal{Z}^+$ & the layer count of the transformer in the VLLM. \\
\midrule
$l_e$ & $[1,2,...,L]$ & the index of layer to be edted. \\
\midrule
$d_h$ & $\mathcal{Z}^+$ & hidden state dimension of the transformer in the VLLM \\
\midrule
$h_n^l$ & $\mathcal{R}^{N\times d_h}$ & \begin{tabular}[c]{@{}l@{}}hidden state of the $n$-th token in the transformer at $l$-th layer. \\ $\tilde{h},\bar{h}$ are the perturbed hidden state and the hidden state of editing sample, respectively. \\ $\dot{h},\ddot{h}$ are the intermediate results in Influence Mapper. \\ $\acute{h},\grave{h},\breve{h}$ are the hidden states of reliability, generality, and locality samples, respectively. \\ $\mathring{h}$ is the hidden state randomly sampled from $\{\acute{h},\grave{h}\}$.\end{tabular} \\
\midrule
$m_n^l$ & $\mathcal{R}^{N\times d_h}$ & MLP output of the $n$-th token in the transformer at $l$-th layer. \\
\midrule
$a_n^l$ & $\mathcal{R}^{N\times d_h}$ & Attention output of the $n$-th token in the transformer at $l$-th layer. \\
\bottomrule
\end{tabular}

\caption{Variable Symbols.}
\label{tab_notations_variable}
\end{table*}

\begin{table}[t!]
    \setlength{\tabcolsep}{3.5pt}
    \renewcommand{\arraystretch}{0.51}

\begin{tabular}{cl}
\toprule
\textbf{Functions} & \textbf{Explanation \& Variants} \\
\midrule
$f_\theta$ & \begin{tabular}[c]{@{}l@{}} VLLM that maps input image-text pair to text \\ response, i.e., $o=f_\theta(x_v, x_t)$. \\ $\hat{f}_\theta$ is the language transformer in the VLLM.\end{tabular} \\
\midrule
$C_{o}(r)$ & contribution of $r$ to $o$ \\
\midrule
$\delta$ & softmax \\
\midrule
$\sigma$ & sigmoid \\
\midrule
$\mathbb{E}$ & expectation \\
\midrule
$\text{ME}$ & model editor \\
\midrule
$\oplus$ & concatenation \\
\bottomrule
\end{tabular}

\caption{Function Symbols.}
\label{tab_notations_function}
\end{table}

\section{Details of Experimental Settings}
\label{Appendix_detail_exp_settings}
\subsection{Datasets:} \textbf{E-VQA} \cite{MMEdit} is the dataset to edit VLLMs to correct error-prone samples in VQA-v2 \cite{DBLP:conf/cvpr/GoyalKSBP17}, including 6,345 training and 2,093 testing samples. 
The VQA task involves presenting a VLLM with an image and a related question, requiring the VLLM to analyze the visual content and the question to provide an accurate textual answer.
\textbf{E-IC} \cite{MMEdit} is the dataset to edit VLLMs to correct error-prone samples in COCO Caption \cite{DBLP:journals/corr/ChenFLVGDZ15}, including 2,849 training and 1,000 testing samples. The IC task involves generating a descriptive textual caption for a given image, which requires the model to understand and articulate the visual content accurately.
Each sample in these two datasets comprises an edit sample, two samples respectively for modal and text generality, and two samples respectively for modal and text locality. For their generality samples, rephrased images and queries are generated using Stable Diffusion \cite{DBLP:conf/cvpr/RombachBLEO22} and ChatGLM \cite{DBLP:conf/acl/DuQLDQY022} respectively. For the locality samples, unrelated images and queries are sourced from the OK-VQA \cite{DBLP:conf/cvpr/MarinoRFM19} and NQ dataset \cite{MEND} datasets, respectively.

\subsection{VLLM Backbones:} 
\textbf{BLIP2} \cite{BLIP2} learns a visual query transformer, called Q-Former, through a two-stage pre-training process to extract the most crucial information from visual data and bridge the representational gap between the frozen visual encoder \cite{DBLP:conf/icml/RadfordKHRGASAM21} and the frozen language model. BLIP2 includes multiple variants, and in this paper, we follow \cite{MMEdit} and choose to experiment with BLIP2-OPT\footnote{\url{https://huggingface.co/Salesforce/blip2-opt-2.7b}}.
Building on BLIP2, \textbf{MiniGPT-4}\footnote{\url{https://huggingface.co/Vision-CAIR/MiniGPT-4}} \cite{MiniGPT-4} freezes the visual encoder, Q-Former, and the language model Vicuna \cite{chiang2023vicuna}, training only a linear layer after the Q-Former to align the visual features with Vicuna.
\textbf{LLaVA}\footnote{\url{https://huggingface.co/liuhaotian/llava-v1.5-7b}} \cite{llava} utilizes GPT-4 \cite{achiam2023gpt} to construct the instruction tuning dataset for VLLM pre-training and achieves the transformation from visual representations to linguistic representations by only training a two-layer MLP between the visual encoder and the language model LLaMA \cite{DBLP:journals/corr/abs-2302-13971}. 
Compared to BLIP2, which compresses visual representations using Q-Former, LLAVA transforms all visual representations, which has the advantage of not losing visual information but the disadvantage of lower inference efficiency.

\subsection{Baseline Editors:} 
Following \citet{MMEdit}, \textbf{FT} (Fine-Tuning) includes two variants, where \textbf{FT-V} fine-tunes the visual encoder module of the VLLM for an edit sample, and \textbf{FT-L} fine-tunes the last layer of the language model. \textbf{KE} \cite{KnowledgeEditor} generates editing offsets for the FFN matrix in the LLM by training an LSTM \cite{DBLP:journals/neco/HochreiterS97} that takes embedded edit samples as inputs. \textbf{MEND} \cite{MEND} trains an MLP to generate the offsets by inputting the decomposed back-propagation gradient of the FFN matrix on an edit sample, thus enhancing editing efficiency. \textbf{IKE} \cite{InContextKnowledgeEdit} constructs demonstrations and uses in-context learning to modify the LLM’s response. \textbf{SERAC} \cite{SERAC} trains a classifier and a counterfactual language model, using the classifier to redirect subsequent inputs related to the edit sample to the counterfactual model for generating responses. \textbf{TP} \cite{T-Patcher} inserts and trains a new FFN layer neuron for an edit sample. \textbf{LTE} \cite{LTE} fine-tunes the LLM to follow editing instructions appended before the inputted query.

\subsection{Model Settings and Training Details:}
\begin{figure*}[t]
\centering
\includegraphics[width=1.\textwidth]{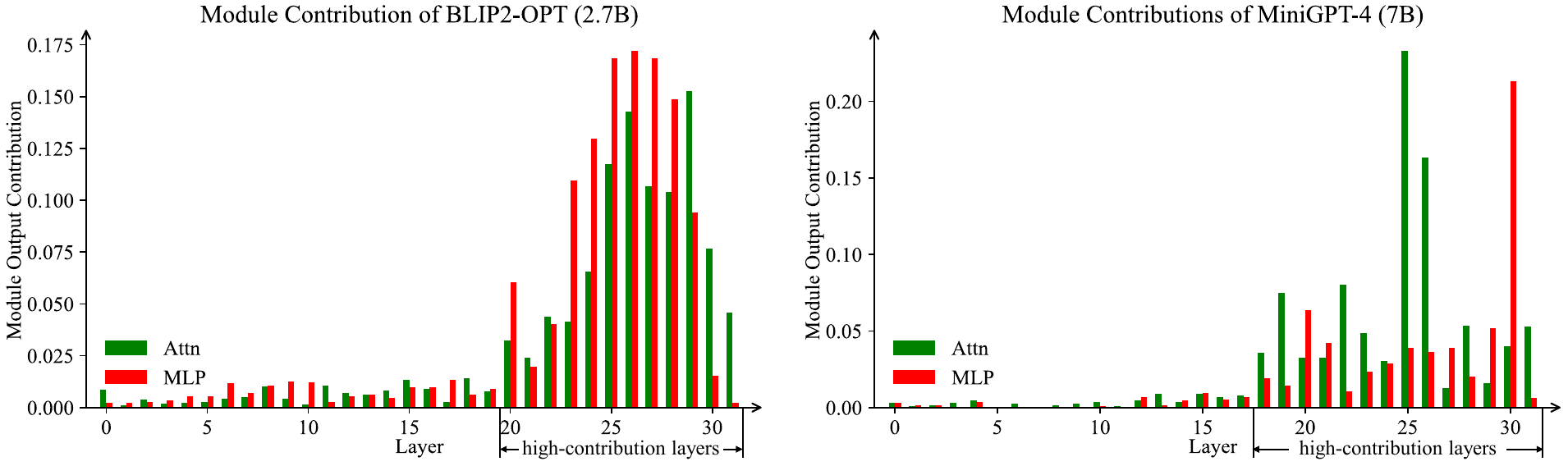} 
\caption{Module contribution of BLIP-OPT (2.7B) and MiniGPT-4 (7B).}
\label{img_module_attribution_blip2_minigpt4}
\end{figure*}

\textbf{For \emph{VisEdit}}, the hyper-parameters of VEAD are set consistently across different backbones.  
Generally, a larger intermediate dimension produces better representational capacity. Constrained by device resources, we set the module dimension to $d_a=1024$.
Similarly, we set $|\mathcal{N}|=24$ to reduce the computational power required for real-time dynamic attribution during training, where a larger value would make IM more stable in approximating the visual attention pattern according to the prompt.
% The loss balance coefficient is set to $\lambda=0.1$. 
% In IM loss, the number of image representation samples in the dynamic attribution experiment is set to $|\mathcal{N}|=24$. 
To effectively exploit the visual information extraction capabilities of VLLM, we insert VEAD before the high-contribution layers. According to Figure \ref{img_attribution} and Figure \ref{img_module_attribution_blip2_minigpt4}, VEAD is inserted at layers 19, 18, and 17 in BLIP-OPT, LLaVA-V1.5, and MiniGPT-4, with training parameters of 21M, 33M, and 33M, respectively.
We set the learning rate to ($\eta=1e-4$), the training batch size to $B=4$, and the maximum number of iterations to 200K. A checkpoint is saved every 500 iterations, and ultimately, the one with the smallest loss is selected for evaluation. The training process requires approximately 2 days on 2 NVIDIA A800 GPUs. These experiments are presented on average with $5$ random runs, using different random seeds but the same hyper-parameters.
\textbf{For the baselines}, the settings of TP \cite{T-Patcher} and LTE \cite{LTE} are referenced from their respective papers. For the other baselines, we follow the same settings as described by \citet{MMEdit} for training and evaluation.

\begin{figure*}[t]
\centering
\includegraphics[width=1.\textwidth]{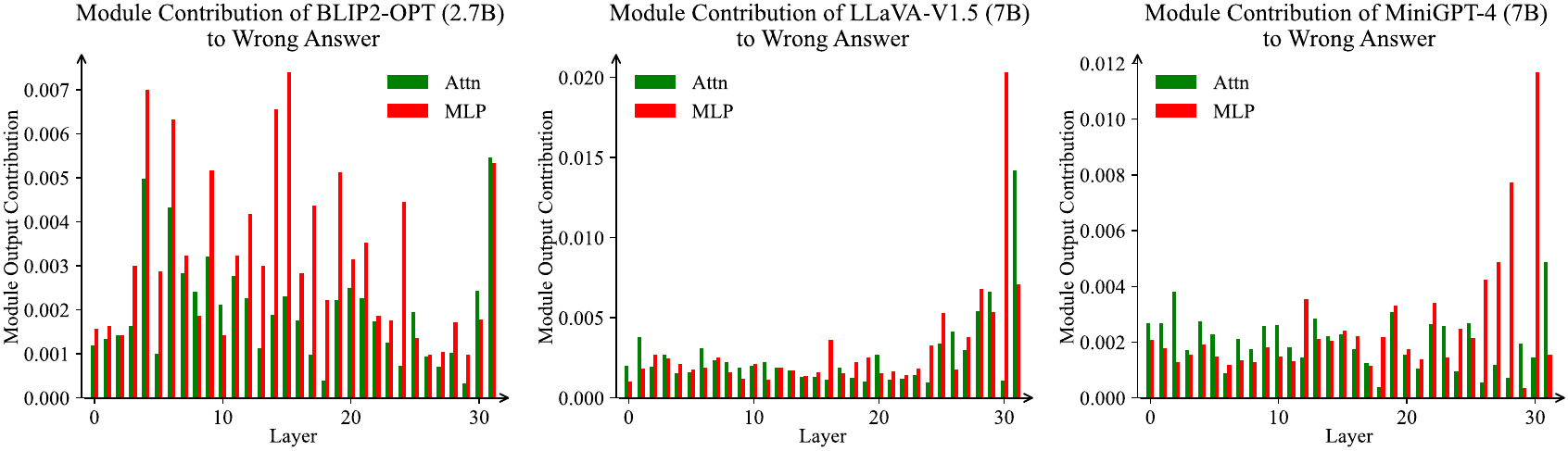} 
\caption{Module contribution of BLIP-OPT (2.7B), LLaVA-V1.5 (7B), and MiniGPT-4 (7B) on wrong key tokens.}
\label{img_module_attritbution_to_wrong_answer}
\end{figure*}

\begin{figure*}[t]
\centering
\includegraphics[width=1.\textwidth]{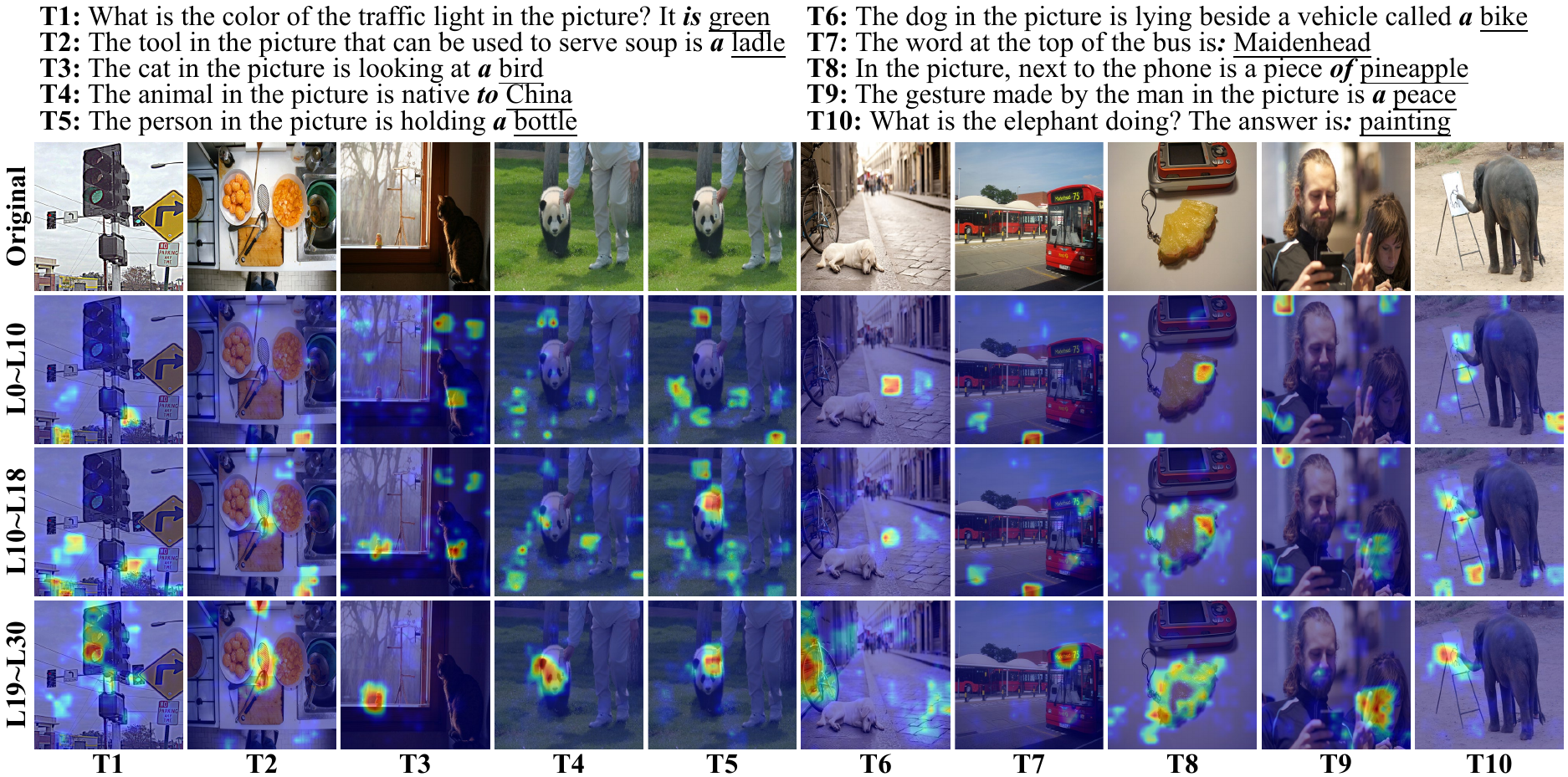} 
\caption{
Visualization of contribution attribution for visual representations, where red indicates higher contributions and blue indicates lower.
\textbf{T*} and \textbf{L*} respectively indicate the test sample index and the attribution analysis layer index of the visual representations. In each sample, the italicized bold text and the underlined text respectively represent the token that predicts the key token and the key token to be predicted.
}
\label{img_more_visual_rep_attribut}
\end{figure*}

\section{Additional Attribution Analysis}
\label{sec_more_attribution_analysis}

\subsection{Module Contribution Attribution}
\noindent\textbf{For BLIP2-OPT and MiniGPT-4:} To verify whether other VLLMs exhibit a similar module contribution distribution across layers as observed in LLaVA-V1.5 \cite{llava} (bar chart in Figure \ref{img_attribution}), we present in Figure \ref{img_module_attribution_blip2_minigpt4} the average attribution results of module contribution for BLIP2-OPT \cite{BLIP2} and MiniGPT-4 \cite{MiniGPT-4}, both of which are evaluated on E-VQA \cite{MMEdit}.
Compared with Figure \ref{img_attribution}, although the contribution value scales and contribution distributions across layers differ slightly among the VLLMs, it can also be concluded that the module outputs in the deeper layers of these two VLLMs contribute more significantly to the final results compared to the shallower layers.

\noindent\textbf{On Wrong Key Tokens:}We conduct module contribution attribution after replacing the key tokens with wrong tokens (the answers of locality samples) to validate the significance of the original attribution results. As shown in Figure \ref{img_module_attritbution_to_wrong_answer}, compared to the original results in Figure \ref{img_attribution} and Figure \ref{img_module_attribution_blip2_minigpt4}, the scale of module contributions decreases by an order of magnitude. Furthermore, the distribution of contributions across layers does not exhibit a significant trend of being low in early layers and high in later layers, except for relatively larger contributions in the last few layers; however, the scale remains small. 
This suggests that the module contribution attribution in our experiment when predicting key tokens is significant and solid.

\subsection{Visual Representation Contribution Attribution}
\label{sec_more_visual_rep_Attribution}
Figure \ref{img_more_visual_rep_attribut} presents more visualization results of visual representation attribution for LLaVA-V1.5 \cite{llava}. 
Overall, in layers 0-10, the relatively important areas visualized typically lack specific semantic information. 
In layers 10-18, the model focuses more intensely on specific areas based on the prompt in some cases; however, the noise remains substantial.
At layers 19-30, it can be observed that the visual areas the model focuses on are comprehensible to humans and highly relevant to the prompt. 
Specifically, T2 and T4 indicate that the model converts its general knowledge into a visual information retriever to focus on specific objects in the image. 
T3, T6, T7, and T8 demonstrate that the model can understand the spatial relationships between objects in the image.
T4 and T5 confirm that the model indeed focuses on different visual areas based on different prompts. T9 and T10 show that the model interprets action semantics in the image to generate responses.

\section{Additional Experimental Results}

\begin{table*}[t!]
\centering

    \footnotesize
    % \scriptsize
    \setlength{\tabcolsep}{2.8pt}
    \renewcommand{\arraystretch}{0.95}

\begin{tabular}{cccccccccccccc}
\toprule
 &  & \multicolumn{6}{c}{\textbf{E-VQA}} & \multicolumn{6}{c}{\textbf{E-IC}} \\
\multirow{-2}{*}{\textbf{Backbone}} & \multirow{-2}{*}{\textbf{Editor}} & \textbf{Rel.} & \textbf{T-Gen.} & \textbf{M-Gen.} & \textbf{T-Loc.} & \textbf{M-Loc.} & \textbf{Average} & \textbf{Rel.} & \textbf{T-Gen.} & \textbf{M-Gen.} & \textbf{T-Loc.} & \textbf{M-Loc.} & \textbf{Average} \\
\midrule
 & FT-V & 27.12 & 22.04 & 21.75 & 100.00 & 87.80 & 51.74$_{(\pm1.02)}$ & 46.69 & 45.58 & 44.02 & 100.00 & 90.85 & 65.43$_{(\pm0.42)}$ \\
 & FT-L & 29.85 & 23.69 & 24.66 & 98.86 & 88.81 & 53.17$_{(\pm1.07)}$ & 46.87 & 45.79 & 44.36 & 99.08 & 92.51 & 65.72$_{(\pm0.99)}$ \\
 & \cellcolor[HTML]{E7E6E6}KE & \cellcolor[HTML]{E7E6E6}87.77 & \cellcolor[HTML]{E7E6E6}86.62 & \cellcolor[HTML]{E7E6E6}3.76 & \cellcolor[HTML]{E7E6E6}97.15 & \cellcolor[HTML]{E7E6E6}55.77 & \cellcolor[HTML]{E7E6E6}66.21$_{(\pm0.00)}$ & \cellcolor[HTML]{E7E6E6}35.10 & \cellcolor[HTML]{E7E6E6}24.20 & \cellcolor[HTML]{E7E6E6}5.89 & \cellcolor[HTML]{E7E6E6}96.78 & \cellcolor[HTML]{E7E6E6}52.22 & \cellcolor[HTML]{E7E6E6}42.84$_{(\pm0.00)}$ \\
 & \cellcolor[HTML]{E7E6E6}IKE & \cellcolor[HTML]{E7E6E6}71.72 & \cellcolor[HTML]{E7E6E6}40.23 & \cellcolor[HTML]{E7E6E6}70.59 & \cellcolor[HTML]{E7E6E6}13.46 & \cellcolor[HTML]{E7E6E6}2.00 & \cellcolor[HTML]{E7E6E6}39.60$_{(\pm0.00)}$ & \cellcolor[HTML]{E7E6E6}68.60 & \cellcolor[HTML]{E7E6E6}59.80 & \cellcolor[HTML]{E7E6E6}63.58 & \cellcolor[HTML]{E7E6E6}12.51 & \cellcolor[HTML]{E7E6E6}2.96 & \cellcolor[HTML]{E7E6E6}41.49$_{(\pm0.00)}$ \\
 & \cellcolor[HTML]{E7E6E6}SERAC & \cellcolor[HTML]{E7E6E6}87.20 & \cellcolor[HTML]{E7E6E6}84.60 & \cellcolor[HTML]{E7E6E6}81.87 & \cellcolor[HTML]{E7E6E6}100.00 & \cellcolor[HTML]{E7E6E6}0.33 & \cellcolor[HTML]{E7E6E6}70.80$_{(\pm0.00)}$ & \cellcolor[HTML]{E7E6E6}40.20 & \cellcolor[HTML]{E7E6E6}36.60 & \cellcolor[HTML]{E7E6E6}35.12 & \cellcolor[HTML]{E7E6E6}100.00 & \cellcolor[HTML]{E7E6E6}0.97 & \cellcolor[HTML]{E7E6E6}42.58$_{(\pm0.00)}$ \\
 & \cellcolor[HTML]{E7E6E6}MEND & \cellcolor[HTML]{E7E6E6}95.51 & \cellcolor[HTML]{E7E6E6}95.27 & \cellcolor[HTML]{E7E6E6}86.37 & \cellcolor[HTML]{E7E6E6}98.73 & \cellcolor[HTML]{E7E6E6}71.33 & \cellcolor[HTML]{E7E6E6}89.44$_{(\pm0.00)}$ & \cellcolor[HTML]{E7E6E6}87.10 & \cellcolor[HTML]{E7E6E6}84.10 & \cellcolor[HTML]{E7E6E6}80.60 & \cellcolor[HTML]{E7E6E6}98.34 & \cellcolor[HTML]{E7E6E6}59.53 & \cellcolor[HTML]{E7E6E6}81.93$_{(\pm0.00)}$ \\
 & TP & 42.96 & 41.53 & 40.70 & 93.32 & 84.61 & 60.62$_{(\pm1.42)}$ & 52.55 & 51.93 & 49.65 & 85.56 & 72.66 & 62.47$_{(\pm0.87)}$ \\
 & LTE & 95.92 & 95.25 & 94.92 & 87.18 & 89.72 & 92.60$_{(\pm1.87)}$ & 89.68 & 87.48 & 86.15 & 86.52 & 87.48 & 87.46$_{(\pm1.06)}$ \\
\multirow{-9}{*}{\begin{tabular}[c]{@{}c@{}}MiniGPT-4\\      (7B)\end{tabular}} & \textit{VisEdit} & \textbf{96.83} & \textbf{96.46} & \textbf{95.38} & \textbf{100.00} & \textbf{90.88} & \textbf{95.91}$_{(\pm1.15)}$ & \textbf{93.25} & \textbf{90.32} & \textbf{89.75} & \textbf{100.00} & \textbf{94.09} & \textbf{93.48}$_{(\pm0.45)}$\\
\bottomrule

\end{tabular}

\caption{Editing performance of MiniGPT-4 on E-VQA and E-IC datasets. ``Rel.'', ``T/M-Gen.'' and ``T/M-Loc.'' stand for reliability, text/modal generality, and text/modal locality, respectively. Results with a gray background are taken from \citet{MMEdit}. The t-tests demonstrate our improvements are statistically significant with $p < 0.05$ level.}

\label{tab_main_exp_extra}
\end{table*}

\subsection{Editing Performance on MiniGPT-4}
\label{sec_minigpt4_editing_performance}
Editing experiments on MiniGPT-4 (7B) are presented in Table \ref{tab_main_exp_extra}. 
The results also show a similar conclusion with general results, demonstrating the efficacy of our method. 
Compared with the editing performance of various editors on LLaVA-V1.5 (7B) \cite{llava} of the same scale in Table \ref{tab_main_exp}, their performance is relatively lower on MiniGPT-4 (7B) \cite{MiniGPT-4}. We hypothesize that this is because LLaVA inputs all visual representations into the language model without compressing the visual information, enabling the editor to more effectively correct the model's descriptions of image details.

\begin{figure*}[t]
\centering
\includegraphics[width=1.\textwidth]{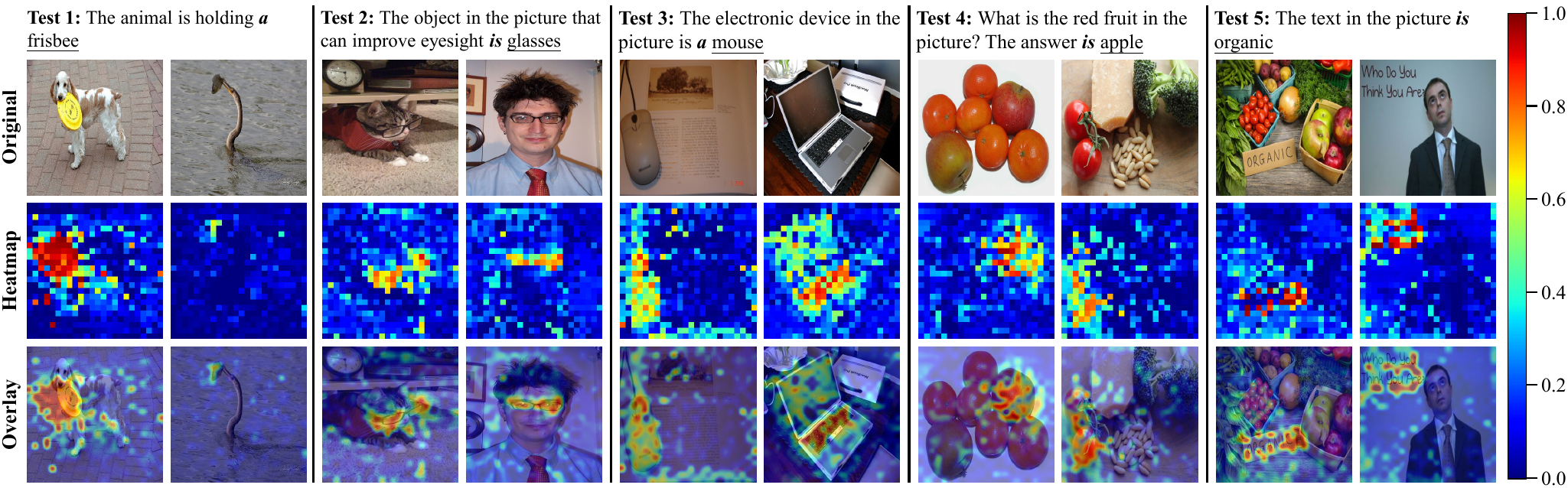} 
\caption{The visualization of IM module in VEAD integrated into LLaVA-V1.5. In each test, VEAD is first edited using the left image along with the prompt. Then, the outputs of IM are visualized for both images.}
\label{img_more_im_visualization}
\end{figure*}

\begin{figure*}[t]
\centering
\includegraphics[width=1.\textwidth]{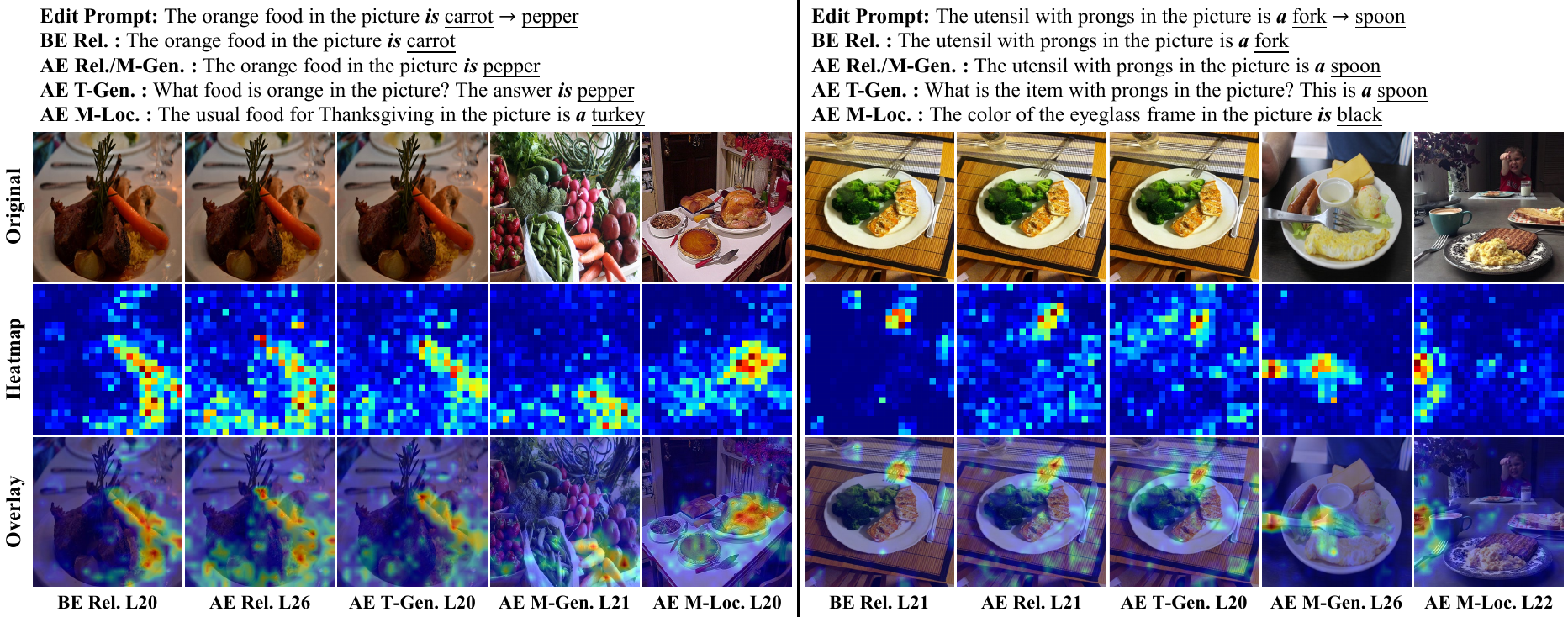} 
\caption{
% Post-\emph{VisEdit}-edited visual representation attribution visualization on two counterfactual edit samples. 
Visualization of visual representation attribution after \emph{VisEdit} edits a counterfactual sample. ``\textbf{BE}'' and ``\textbf{AE}'' indicate Before Editing and After Editing respectively. \textbf{L*} indicates the layer index. }
\label{img_more_vead_post_edit_visualization}
\end{figure*}

\subsection{Results of IM Visualization}
\label{sec_additional_im_visualization}
Figure \ref{img_more_im_visualization} shows more visualization results of the IM outputs.
The results demonstrate that IM indeed learns the attention patterns that use the last token of the edit prompt, integrating the overall prompt information by the VLLM, to focus on the relevant visual representations.

\subsection{Instance Analysis for VEAD}
\label{sec_additional_instance_analysis_VEAD}
Figure \ref{img_more_vead_post_edit_visualization} displays additional visualization results of visual representation attributions using VEAD to edit counterfactual samples. It can be observed that the editing process of VEAD does not significantly change the highly contributing visual areas, but the model responses are changed on reliability and generality samples. 
This indicates that for reliability and generality samples, the corresponding visual representations are edited into those representations that highly contribute to predicting the edited token (e.g., ``pepper'' in the left edit sample).

\end{document}